\documentclass[preprint,12pt]{elsarticle}

\usepackage{cite}
\usepackage{float}
\usepackage{arydshln}
\usepackage{makecell}
\usepackage{balance}
\usepackage{graphics} 
\usepackage{epsfig} 
\usepackage{mathptmx} 
\usepackage{times} 
\usepackage{amsmath} 
\usepackage{amssymb}  
\usepackage{xcolor}
\usepackage{colortbl}
\usepackage{algorithm2e}
\usepackage{algpseudocode}
\usepackage{tabularx}
\usepackage{caption}
\usepackage[breaklinks,colorlinks]{hyperref}
\usepackage{caption}
\usepackage{subcaption}
\usepackage{multirow}

\usepackage{enumitem}
\usepackage{booktabs}
\usepackage{physics}
\usepackage[normalem]{ulem}

\definecolor{mygreen}{rgb}{0.0, 0.5, 0.0}
\usepackage{tabularx, booktabs}
\usepackage{cleveref}
\usepackage{fancyvrb}
\usepackage{amsfonts}
\usepackage{balance}

\usepackage{amssymb}
\usepackage{amsmath}

\usepackage{lineno}
\newcommand{\formattedparagraph}[1]{\noindent \textbf{#1}}

\newcommand{\revised}[1]{\textcolor{black}{ #1}}
\newcommand{\revisedtwo}[1]{\textcolor{black}{ #1}}

\journal{Journal of Photogrammetry and Remote Sensing}

\begin{document}

\begin{frontmatter}



\title{Mobile Robotic Multi-View Photometric Stereo}


\author{Suryansh Kumar} 

\affiliation{organization={Visual Computing and Computational Media Section (VCCM), College of Performance Visualization and Fine Arts (PVFA), Department of Electrical and Computer Engineering (ECEN), Department of Computer Science and Engineering (CSCE)},
            addressline={LAAH Building},
            city={College Station, Texas A\&M University},
            postcode={77843}, 
            state={Texas},
            country={United States of America.}
            }

\begin{abstract}
Multi-View Photometric Stereo (MVPS) is a popular method for fine-detailed 3D acquisition of an object from images. Despite its outstanding results on diverse material objects, a typical MVPS experimental setup requires a well-calibrated light source and a monocular camera installed on an immovable base. This restricts the use of MVPS on a movable platform, limiting us from taking MVPS benefits in 3D acquisition for mobile robotics applications. To this end, we introduce a new mobile robotic system for MVPS. While the proposed system brings advantages, it introduces additional algorithmic challenges. Addressing them, in this paper, we further propose an incremental approach for mobile robotic MVPS. Our approach leverages a supervised learning setup to predict per-view surface normal, object depth, and per-pixel uncertainty in model-predicted results. A refined depth map per view is obtained by solving an MVPS-driven optimization problem proposed in this paper. Later, we fuse the refined depth map while tracking the camera pose w.r.t the reference frame to recover globally consistent object 3D geometry. Experimental results show the advantages of our robotic system and algorithm, featuring the local high-frequency surface detail recovery with globally consistent object shape. Our work is beyond any MVPS system yet presented, providing encouraging results on objects with unknown reflectance properties using fewer frames without a tiring calibration and installation process, enabling \revised{computationally efficient} robotic automation approach to photogrammetry. \revisedtwo{The proposed approach is nearly 100 times computationally faster than the state-of-the-art MVPS methods such as \citep{kaya2023multi, kaya2022uncertainty} while maintaining the similar results when tested on subjects taken from the benchmark DiLiGenT MV dataset \citep{li2020multi}. Furthermore, our system and accompanying algorithm is data-efficient, i.e., it uses significantly fewer frames at test time to perform 3D acquisition\footnote{The subjects used to train, test, and compare the results can be downloaded from \href{https://sites.google.com/site/photometricstereodata/mv}{here}.}.}
\end{abstract}

    
\end{frontmatter}


\section{Introduction}\label{sec:intro}
In recent years, 3D data acquisition of objects from images has become increasingly popular due to its high demand from industries involved in spatial computing. These industries require high-quality 3D data to train large deep-learning models to accomplish their computing goals. Although methods and apparatus exist to solve the task, they proved to be costly---both manually and economically. Hence, one key goal yet to be achieved is a fully automated mobile robotic system that can provide high-quality 3D data of objects from images. To this end, we propose the first portable ``Mobile Robotic Multi-view Photometric Stereo'' system and supporting algorithm to perform 3D acquisition.

Over the years, methods such as structure from motion (SfM) \citep{schoenberger2016sfm}, multiview stereo (MVS) \citep{furukawa2015multi}, photometric stereo (PS) \citep{woodham1980photometric}, multiview photometric stereo (MVPS) \citep{hernandez2008multiview}, and more recently, neural radiance fields (NeRF) \citep{mildenhall2020nerf}, 3D Gaussian Splatting \citep{kerbl3Dgaussians} and their variations \citep{ mueller2022instant, jain2024learning} have emerged as some of the popular practical approaches in 3D data acquisition from images. Yet, MVPS stands out when it comes to the precision, details, and accuracy. One primary reason is that PS images in MVPS setup help recover an object's high-frequency local details, while MVS images help preserve the object's global structural content. \revisedtwo{Nehab et al. \citep{nehab2005efficiently} is one the early approaches to practically demonstrate this idea of exploiting complementary nature of PS and MVS, which is recently exploited using deep-learning approaches by Kaya et al. \citep{kaya2022uncertainty, kaya2023multi}}. This powerful idea makes MVPS as a default method of choice in forensics \citep{moons20093d}, metrology \citep{nehab2005efficiently}, archaeology \citep{chatterjee2015photometric, kaya2022uncertainty} and other scientific as well as engineering disciplines \citep{bronstein2008numerical}.

\begin{figure*}[t]
    \centering
\includegraphics[width=1.0\textwidth]{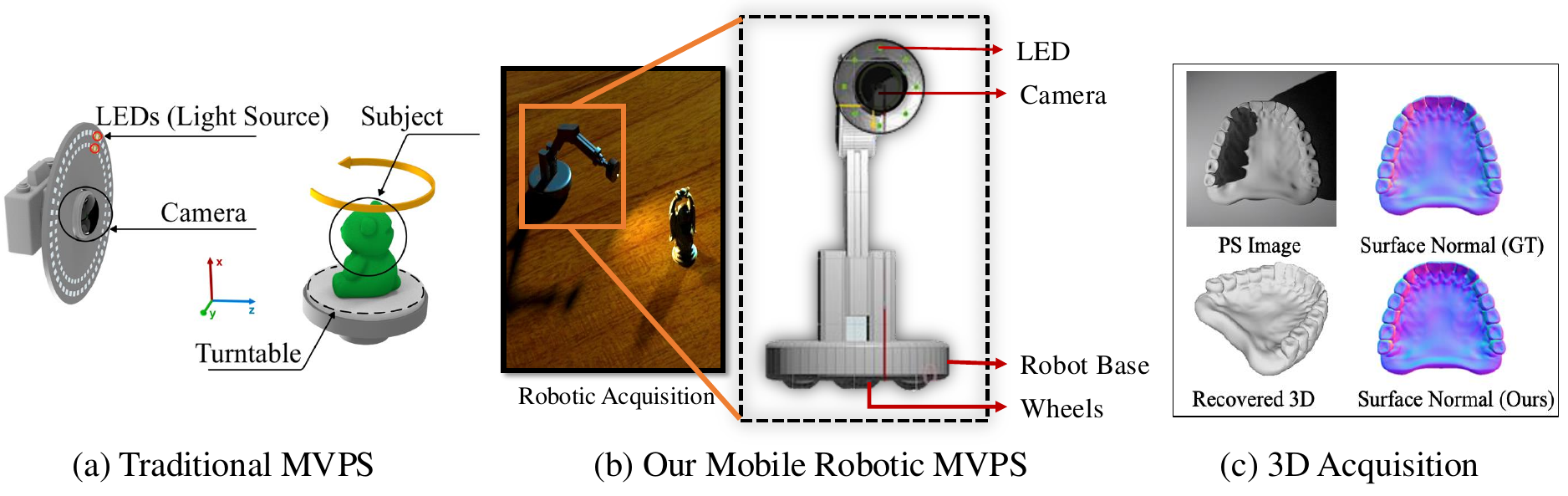}
    \caption{\textbf{(a)} Popular multiview photometric stereo setup as shown in \citep{kaya2021neural}, where the hardware is installed on an immovable base with a subject placed on a turn table. \textbf{(b)} Our mobile robotic MVPS setup, where the robot is allowed to move for object's 3D acquisition. \textbf{(c)} 3D acquisition results using images of tooth model acquired using our system. }
    \label{fig:mvps_hardware_comparison}
\end{figure*}

Despite the apparent advantage of MVPS over other popular approaches, its usage is limited to a controlled lab setup or a industrial machine vision setup. This is mainly due to the MVPS experimental design choice, where an object is placed at the center of a rotating table, while the camera and LEDs are placed at a fixed distance from the table center---refer Fig. \ref{fig:mvps_hardware_comparison}(a). There are many real-world cases where it is difficult to use a classical MVPS setup for 3D acquisition. For example, in fine-grained object geometry, its texture, and material acquisition, where details are visible at varying distance from the object. This is widely witnessed in stones recovered in a planetary exploration or in studying biological organs, etc. Thus, we need a mobile robotic MVPS system for all such cases, \revised{where the MVPS acquisition setup is allowed to move closer or father from the object freely, depending on the object’s size and application requirements.} To our knowledge, this is one of the early attempts in this direction for automating MVPS system.

Our mobile robotic system comprises 8 LED light sources and a monocular camera installed on a mobile robotic arm. The robot arm's end-effector is equipped with LEDs circularly arranged centered around the camera---refer Fig. \ref{fig:mvps_hardware_comparison}(b)\footnote{kindly refer to the appendix for precise hardware details.}. While assembling or simulating our hardware design is relatively easy, executing \textbf{MVPS working principle} on a mobile platform brings additional algorithmic challenges. Unlike classical MVPS, where all MVS and PS images are used to recover camera poses, estimate LEDs light directions, and perform 3D data acquisition, i.e.,  \emph{an offline global approach}, here, we have access to only 8 PS images in a given time from one viewpoint during inference. And since we aim for a mobile robotic platform, it makes more sense to have an \emph{an online incremental approach}, which brings additional challenges in light calibration, camera pose estimation, and noisy or partial shape registration, thereby introducing extra difficulty in detailed 3D data acquisition. This brings us to propose an incremental uncalibrated approach to MVPS. Accordingly, we introduce suitable deep neural networks to infer light calibration parameters, surface normal, uncertainty in surface normal prediction, and depth prior per view at test time. These predicted data help our system refine depth map per view via an uncertainty-based optimization while additionally help in computing robust camera pose parameters of live frame w.r.t the reference frame. Later, refined depth map per view is fused using an online fusion approach for detailed object's 3D recovery (see Fig. \ref{fig:mvps_hardware_comparison}(c)). In summary, this paper makes the following contributions.

\begin{itemize}[noitemsep]
    \item \textbf{Automating MVPS-Based Photogrammetry}: This work introduce a novel mobile robotic \revised{MVPS} system, supported by a custom algorithm, aimed at achieving dense, detailed 3D data acquisition of an object. This represents an initial step toward automating MVPS based photogrammetry system.
    
    \item \textbf{Incremental Online MVPS Approach}: Unlike traditional MVPS techniques, we propose an incremental, online methodology that aligns with our mobile robotic hardware. This approach enhances efficiency and adaptability in 3D data acquisition processes depending on object's surface details.
    
    \item \textbf{Uncertainty-Driven Optimization}: We present an uncertainty-driven optimization tailored for an incremental fusion pipeline, significantly improving the quality of depth maps. Our robotic system yields high-qualtity results with fewer images compared to the state-of-the-art MVPS methods.

    \item \revised{\textbf{Adaptability to Object Detail}: The mobile nature of the system allows for flexible positioning of the MVPS setup, enabling capture of object data at different distance from the object to capture surface details, which is not feasible with traditional static setups.} 
\end{itemize}

\revised{The proposed method contribution brings the following advantage compared to the existing MVPS system(s):
\begin{itemize}[noitemsep]
    \item \textbf{Data efficiency}:  Our method uses only 36 views × 8 images per view = 288 images compared to other methods which uses 1920 images in total to provide results with a similar accuracy.
    \item \textbf{Computational efficiency}: Our method is more than 100x computationally faster than the state-of-the-art MVPS method(s) while achieving similar 3D reconstruction accuracy results.
    \item \textbf{Cost efficiency}: Commercially available MVPS systems for high-quality 3D data acquisition are costly. For instance, Arago MVPS photogrammetry RIG system typically ranges between \$10,000 and \$20,000 USD; again, this is a static MVPS setup. On the contrary, our mobile MVPS system can be built for less than \$2,500 USD due to simple and portable design.
\end{itemize}
}

\section{Related Works}\label{sec:relatedworks}
Hern${\acute{\text{a}}}$ndez et al. \citep{hernandez2008multiview} introduced the classical MVPS setup. Yet, Nehab et al. \citep{nehab2005efficiently} pioneered the integration of PS images with active range scanning techniques, highlighting the complementary nature of shape from shading and active range modalities in 3D data acquisition. However, the widely used MVPS setup involves a turntable experimental setup, where PS images of the subject positioned on the turntable are captured from a staged viewpoint. It is essential to highlight that within this setup, both the camera and the lighting sources maintain a stationary position, with only the turntable undergoing rotation. This rotation facilitates the acquisition of the subject's viewpoints from different angles with each turn. Specifically, each turntable rotation facilitates capturing and storing MVS and PS images for each lighting source (see Fig.\ref{fig:mvps_hardware_comparison}(a)). 

\vspace{-0.3cm}
\revisedtwo{
\subsection{Traditional MVPS Methods}
}
Earlier MVPS approaches often relied on a specific Bidirectional Reflectance Distribution Function (BRDF) model, leading to unreliable outcomes for real-world objects whose reflectance properties deviate from the presupposed BRDF model. To this end, \citep{park2016robust} introduced a method based on piece-wise planar mesh parameterization, designed to enhance the object's fine surface details reconstruction through displacement texture maps. Yet, their work overlooked surface reflectance modeling. Contrarily, other methodologies, such as \citep{ren2011pocket, dong2010manifold}, do engage in BRDF modeling. Still, their usage is limited to nearly flat surfaces and presupposes a prior knowledge of the surface normals.

\vspace{-0.2cm}
\revisedtwo{
\subsection{Deep Learning based MVPS Methods}
}
In recent years, deep learning methodologies have been suggested as better alternatives to traditional MVPS techniques, albeit employing classical MVPS hardware setups. To this end, \citep{zhang2022iron} proposed a neural inverse rendering approach to reconstruct objects' shape and material attributes. However, this method is predicated on the assumption of a co-located camera and lighting source, hence incompatible with standard MVPS setups. On the contrary, \citep{kaya2021neural}  proposed a method based on neural radiance fields. This approach endeavors to estimate the object's surface normal and integrates them within a volume rendering formula to enhance object representation learning. Despite its conceptual benefit, this technique struggles to accurately capture objects' high-quality geometric details.
Conversely, an uncertainty-based MVPS methodology has been introduced \citep{kaya2022uncertainty}, demonstrating superior performance in 3D object reconstruction. Recently, \citep{kaya2023multi} extended \citep{kaya2022uncertainty} to make deep MVPS work for diverse material types. Whereas, \citep{zhao2023mvpsnet} worked on speeding up the deep-MVSP model inference time while maintaining the 3D acquisition accuracy. Not long ago, \citep{lichy2021shape} proposed estimating the geometry and reflectance of objects using a camera, flashlight, and a tripod setup. Whereas \citep{Cheng_2023_CVPR} uses a smartphone’s built-in flashlight and combines darkroom photometric methods with neural light fields \citep{wang2021neus} to recover an object's 3D geometry. 

\begin{figure*}[t]
    \centering
\includegraphics[width=1.0\textwidth]{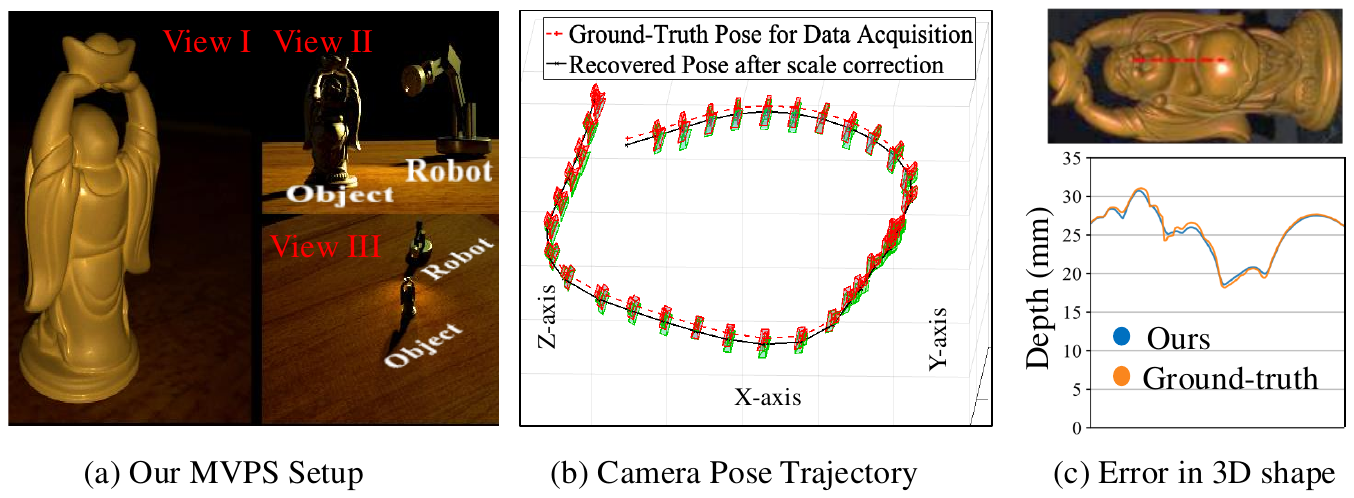}
    \caption{\textbf{Our mobile robotic test time setup.} (a) Our mobile robot moves around the object at test time, performing 3D data acquisition. (b) The robot's ground truth and recovered camera pose trajectory are shown in red and green, respectively. (c) side view of the recovered 3D data compared to ground truth shown in millimeters (mm) along chosen geodesic (shown with a red line on BUDDHA image).}
    \label{fig:img_acquisition_camera_traj}
\end{figure*}

\smallskip
\smallskip
\noindent
Alternatively, to all the approaches to MVPS mentioned above, this paper introduces a novel mobile robotic hardware setup and supporting algorithm based on the MVPS principle for performing an object's 3D acquisition in an incremental setting. Our approach helps recover fine, detailed 3D object geometry under limited lighting conditions. It further provides flexibility in the movement of the MVPS system via a controlled robotic setup for far or near capture, depending on the object size and its surface details, which is not practically possible with previous MVPS experimental setup.

\section{Our Mobile Robotic MVPS}\label{sec:methods}

\subsection{Robotic Acquisition Setup}
As shown in Fig. \ref{fig:mvps_hardware_comparison}(b), our robotic arm's end-effector is equipped with an array of 8 LED light sources, arranged in a circular configuration around a monocular camera, ensuring equidistant positioning from the camera's center. This setup is designed to provide consistent illumination. The camera is fitted with a lens with a focal length of 50mm, which is helpful in high-quality image capture. To facilitate mobility and versatility in data acquisition, the baseplate of the robotic arm is mounted on a robust wheelbase.

Our mobile robotic system presents a distinct contrast to the static setup introduced in the DiLiGenT-MV dataset's acquisition \citep{shi2016benchmark}, which utilizes a stationary MVPS setup with subjects positioned on an turntable. In our robotic MVPS setting, PS imagery is captured from a moving camera with the subjects remaining stationary throughout. At the test time, the robot maneuvers along the given trajectories, attaining a total of 36 distinct poses with an angular separation of 10 degrees between each pose. For each pose, 8 PS images are captured, corresponding to the illumination from each LED, resulting in a processing of $36 \times 8 = 288$ images per object.

Fig.~\ref{fig:img_acquisition_camera_traj}(a) shows the robotic MVPS acquisition setup from three distinct views, whereas Fig.~\ref{fig:img_acquisition_camera_traj}(b), Fig.~\ref{fig:img_acquisition_camera_traj}(c) shows the robot's camera pose trajectory used at test time depicting 36 camera poses and the 3D reconstruction error profile along the chosen geodesic, respectively. To ensure thorough testing of our approach, detailed logs of camera data—encompassing both positions and orientations—as well as the LEDs light direction and intensity are stored. For  experimentation and result comparisons with MVPS methods, we used the 3D objects from DiLiGenT-MV dataset \citep{li2020multi}, which is often characterized by their varied BRDFs and complex surface textures.

\subsection{Proposed Methodology}

\formattedparagraph{Key Notations and MVPS Setup.}
We denote $L$ as the total number of lights and $T$ as the total number of viewpoints for which the robot stops to capture the images. We define $\mathcal{I}_\textrm{ps}^{t} = \{I_1^t, I_2^t, ....., I_L^t\}$ as the set of photometric stereo images captured from a viewpoint at time $t \in [1, T]$. Let's assume $\mathbf{R}^{t} \in {SO(3)}, \lambda\hat{\mathbf{t}}^{t} \in \mathbb{R}^{3 \times 1}$ as the robot camera pose, and $l_{k}^{t} \in \mathbb{R}^{3 \times 1}$, $e_{k}^{t} \in \mathbb{R}_{+}$ as the $k^{th}$ light source direction and corresponding intensity at time $t$. Since, we have $L$ number of images at every instance the robot stops and capture the images, we compute ${I}^{t}_\textrm{si} = \textrm{median}(\mathcal{I}_\textrm{ps}^{t})$ as a single image representative of overall radiosity at time $t$, following \citep{li2020multi} assumption. Such a robotic experimental setup is designed to imitate the MVPS working principle \citep{nehab2005efficiently, hernandez2008multiview}, i.e., estimate surface normal using light-varying PS images, which is excellent at recovering high-frequency surface details. While, ${I}^{t}_\textrm{si}$ help recover object's global surface profile via single image depth prediction model, thereby helping us overcome the low-frequency bias due to PS in surface reconstruction. In our experiments, we assume intrinsic camera calibration matrix $\mathbf{K} \in \mathbb{R}^{3 \times 3}$ is known.

\smallskip
\formattedparagraph{\textit{(i)} Uncalibrated Photometric Stereo.}  For our mobile robotic MVPS system, we aim to infer all the LED's light source direction and intensity followed by the object's surface normal estimation on the fly. And therefore, an uncalibrated approach to PS seems like a reasonable choice \citep{chen2019self, kaya2021uncalibrated}. Accordingly, we train the light estimation network in a supervised setting to infer reliable light source direction and intensity at test time. For surface normal prediction, we train the convolutional neural network-based PS network \citep{ikehata2018cnn}, again in a supervised setting. As is known, perfect prediction of surface normal is challenging; therefore, we leverage the Bayesian uncertainty modeling in our normal estimation network by incorporating drop-out approximation \citep{gal2015bayesian}. The uncertainty modeling in normal estimation network greatly helps us in depth map refinement for effective fusion of reliable 3D data over frames.

We symbolize $\mathcal{X}_\textrm{ps} \in \mathbb{R}^{h \times w \times 3n}$ as all $n$ PS images at train time with $(l_{k}, e_{k}) ~\forall ~k \in [1, n]$ as the light source direction and corresponding intensity pair. Here, $h \times w$ denotes image height and width, respectively. For light source data prediction, we train the deep neural network design proposed in \citep{chen2019self}, and for surface normal prediction, we train the deep PS model proposed in \citep{ikehata2018cnn} with uncertainty modeling based on \citep{kaya2022uncertainty}. At test time, we infer light direction $l_{k}^{t}$, its intensity $e_{k}^{t}$, surface normal map $N_\textrm{ps}^{t} \in \mathbb{R}^{h \times w}$ and per pixel confidence $\Lambda_{ps}^{t} \in \mathbb{R}^{h \times w}$ in surface normal prediction from a given view point at time $t$. Note that we assume the object's 2D image mask is known from each view point, and as a result, we discard pixels outside the object's mask in our experiment and assign 0 confidence value to those pixels with $\mathbf{0}_{3 \times 1}$ as its surface normal value.

\smallskip
\formattedparagraph{\textit{(ii)} Single Image Depth Prediction.} To replicate the operational principles of MVPS in our mobile robotic MVPS system, it becomes imperative to incorporate some form of range data at test time. This necessity arises primarily due to the inherent susceptibility of photometric stereo (PS) techniques to low-frequency biases in 3D reconstruction. The objective is to mitigate such a limitation with PS through the utilization of range data. However, the task of obtaining depth from single image presents significant geometric challenges. Recent advancements in the domain of supervised deep learning, particularly through the application of vision transformer based deep learning models, have shown outstanding results in depth prediction \citep{ranftl2020towards, ranftl2021vision, birkl2023midas, liu2023single, liu2023va}. At present, it is feasible to infer reliable scene depth up to scale using merely a single image input. These breakthroughs in single image depth prediction made our mobile robotic MVPS working possible.

Among all the tested single image depth prediction (SIDP) models, MiDaS v3.1 \citep{birkl2023midas} suitably fits our requirements. The models in MiDaS v3.1 provides high-quality depth prediction result at low compute requirement for real-time applications, thereby providing us with the flexibility in performance optimization based on hardware capabilities. By effectively capturing global image context, MiDaS v3.1 streamlined the depth estimation process and reduced computational overhead. This improved runtime efficiency is crucial for our mobile robotic MVPS system, enabling faster and more responsive depth predictions per view.

Our MVPS system at test time predict depth $D_\textrm{pd}^{t} \in \mathbb{R}^{h \times w}$ by providing ${I}^{t}_\textrm{si}$ to MiDaS v3.1 pre-trained model.

\smallskip
\formattedparagraph{\textit{(iii)} Depth Map Refinement.}
The depth maps generated per view using MiDaS v3.1 demonstrate commendable accuracy in global depth estimation yet exhibit limitations in capturing the finer local surface details, such as scratches and indentations. These limitations present significant challenges to achieving our objective of reconstructing intricate surface details. Adopting PS techniques for measuring object normal fields offers a pathway to reconstructing surfaces with great precision in capturing local details. Yet, surface normal recovered via PS are susceptible to introducing low-frequency biases in 3D reconstructions, which can compromise the integrity of overall depth measurements.

To address the above mentioned limitation, our approach integrates depth map with surface normal information, leveraging the complementary strengths of PS and SIDP for enhanced 3D acquisition of objects. This integration is achieved by solving an optimization function explicitly designed for this purpose. The optimization function proposed herein utilizes the predicted surface normal alongside their associated uncertainties, combining them with SIDP-derived per-view depth data to facilitate the recovery of a high-quality depth map. The devised solution imposes penalties on deviations from the surface normal ascertained through PS, particularly in instances of high-confidence predictions, and applies corresponding adjustments in the inverse scenario. This methodology underscores a strategic effort to balance the influences of both PS and SIDP, thereby improving the resultant 3D acquisition.

The goal is to refine the depth map based on the surface normal prior and depth prior known to us at time $t$. Utilizing the relation between discrete depth map and surface normal via standard finite differences, i.e., the gradient of depth map must be equivalent to surface normal. Denoting $D^{t}$ as the depth map at time $t$, we write $\nabla D^{t} \equiv N_\textrm{ps}^{t}$, with $\nabla$ as the first order gradient operator. Putting this approximation as an optimization cost function, we obtain
\begin{equation}\label{eq:surface_normal_constraint}
\begin{aligned}
& \displaystyle \underset{D^{t}} {\fontfamily{cmtt}\selectfont {\text{minimize}}} 
~ \frac{1}{2}\|(\nabla D^{t} - N_\textrm{ps}^{t})\|^2.
\end{aligned}
\end{equation}
Next, we aim to use the depth prior from the SIDP model, i.e., the refined depth map must conform to SIDP predicted depth in the low-frequency domain, i.e., $D^{t} \equiv D_\textrm{pd}^{t}$, resulting in the following cost function for optimization.
\begin{equation}\label{eq:depth_contraint_per_view}
\begin{aligned}
& \displaystyle \underset{D^{t}} {\fontfamily{cmtt}\selectfont {\text{minimize}}} 
~ \frac{1}{2}\|(\nabla D^{t} - N_\textrm{ps}^{t})\|^2 + \frac{1}{2}\|(D^{t} - D_\textrm{pd}^{t})\|^2.
\end{aligned}
\end{equation}
We can obtain a refined depth map by solving the Eq.\eqref{eq:depth_contraint_per_view}. Yet, Eq.\eqref{eq:depth_contraint_per_view} miss to account for the model prediction uncertainty. Assume  $\Lambda_\textrm{ps}^{t} ~\mathbb{R}^{h \times w}$ as the uncertainty matrix containing per pixel surface normal prediction confidence value. Using $\Lambda_\textrm{ps}^{t}$, we can improve our optimization formulation to cope up with possible error due to model prediction error. Combining Eq.\eqref{eq:surface_normal_constraint} and Eq.\eqref{eq:depth_contraint_per_view} with the model uncertainty in a mindful way, we arrive at our overall depth map refinement optimization, i.e.,
\begin{equation}\label{eq:overall_optimization_per_view}
\begin{aligned}
& \displaystyle \underset{D^{t}_c} {\fontfamily{cmtt}\selectfont {\text{min.}}} 
~~\frac{1}{2}\|\Lambda_{ps}^{t}\odot(\nabla D^{t}_{c} - N_\textrm{ps}^{t})\|^2 + \frac{1}{2}\|(1-\Lambda_{ps}^{t})\odot(D^{t}_c - D_\textrm{pd}^{t})\|^2,
\end{aligned}
\end{equation}
where, $D_{c}^{t}$ denotes the corrected or refined depth map obtained after optimization. $\odot$ in Eq.\eqref{eq:overall_optimization_per_view} denotes the point-wise product popularly known as Hadamard product. Eq.\eqref{eq:overall_optimization_per_view} optimization is easy to understand, i.e.,  a modality confidence-based weighted optimization favoring the suitable modality priors during optimization. More regularization constraints such as Laplacian smoothness \citep{shi2014photometric}, total variational constraint \citep{antensteiner2018variational}, etc. can be added, however, it is observed to make  Eq.\eqref{eq:overall_optimization_per_view} computationally expensive to optimize, hence we avoided them for our robotic setup. As we will see later, Eq.\eqref{eq:overall_optimization_per_view} does a good job in detailed recovery of the object geometry at the same time it can be optimized efficiently using fast gradient based methods \citep{hager2013limited, liu1989limited, berahas2016multi}.

\begin{figure*}[t]
    \centering
\includegraphics[width=1.0\textwidth]{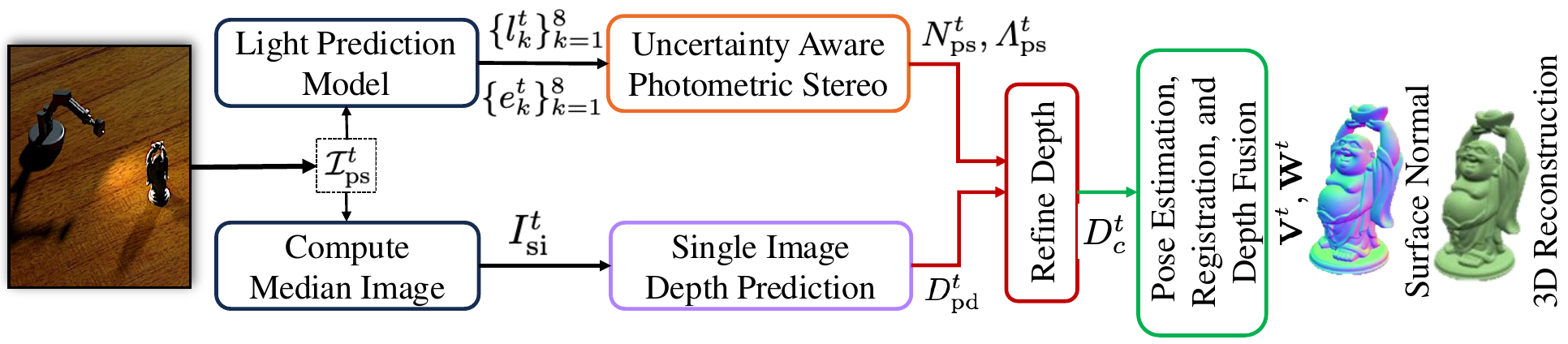}
    \caption{\textbf{System Overview}. Our robotic MVPS setup captures $\mathcal{I}_\textrm{ps}^{t}$ and infers all the light source direction $(\{l_{k}^{t}\}_{k=1}^8)$ and intensity $(\{e_{k}^{t}\}_{k=1}^8)$ followed by ${I}^{t}_\textrm{si}$ computation for that view-point. The recovered light source parameters are then used to predict surface normal map $N_\textrm{ps}^t$ and associated uncertainty $\Lambda_\textrm{ps}^t$ with the predicted values. In parallel, the depth map corresponding to ${I}^{t}_\textrm{si}$ is predicted using the pre-trained SIDP model. Later, a refined depth map is computed by solving Eq.\eqref{eq:overall_optimization_per_view}, which is fused to the global TSDF volume representation over frames for object 3D acquisition.}
    \label{fig:overall_pipeline}
\end{figure*}

\smallskip
\formattedparagraph{\textit{(iv)} Depth Map Fusion over Frames.} The successive refined depth frame obtained after solving Eq.\eqref{eq:overall_optimization_per_view} is fused in an incremental manner, where the local live camera pose and $D_{c}^{t}$ is registered to the reference frame before fusing into a global 3D volumetric representation. This is often termed as Truncated Signed Distance Function (TSDF) fusion in computer graphics literature \citep{curless1996volumetric}. The core idea is to represent the object's 3D using a voxel grid, where each voxel stores a signed distance value indicating the voxel's distance to the nearest surface boundary. Positive values represent distances outside the surface, negative values represent distances inside the surface, and a value of zero indicates the surface itself. The ``truncated'' aspect means that only distances within a certain range around the surface are stored, simplifying the calculations and storage needs.

Our refined depth data fusion involves few steps, including the transformation of local refined depth map into the reference coordinate frame of the TSDF volume $\mathbf{V}^{t}$, the computation of the TSDF value for each voxel based on the new refined depth data, and the update of the voxel values in the volume. Given $D_{c}^{t}$ we extract a local camera-aligned voxel grid from previous time frame with TSDF data $\mathbf{V}^{t-1}_{local}$ and weight $\mathbf{W}^{t-1}_{local}$ from the global voxel grid data with TSDF value $\mathbf{V}^{t-1}$ and weight $\mathbf{W}^{t-1}$. Our fusion pipeline takes $D_{c}^{t}$ along with $\mathbf{V}^{t-1}_{local}$, and $\mathbf{W}^{t-1}_{local}$ to compute the local TSDF update $\mathbf{v}^{t}_{local}$ as well as estimate the camera pose parameters via registration of depth data w.r.t. reference frame \citep{yang2020teaser}. The estimated TSDF update is transferred back into the reference frame to obtain $\mathbf{v}^{t}$ which is then integrated into the global TSDF volumes via well-known TSDF update equation, i.e.,
\begin{equation}
    \mathbf{V}^{t}(\mathbf{p}) = \frac{\mathbf{W}^{t-1}(\mathbf{p})\cdot\mathbf{V}^{t-1}(\mathbf{p}) + \mathbf{w}^{t}(\mathbf{p})\cdot\mathbf{v}^{t}(\mathbf{p})}{\mathbf{W}^{t-1}(\mathbf{p}) + \mathbf{w}^{t}(\mathbf{p})},
\end{equation}

\begin{equation}
    \mathbf{W}^{t}(\mathbf{p}) = \mathbf{W}^{t-1}(\mathbf{p}) + \mathbf{w}^{t}(\mathbf{p}),
\end{equation}
where, $\mathbf{p}$ is a 3D point corresponding to the object. Here we slightly abused our notation by appending brackets and point variable. The signed distance update $\textbf{v}^t(\textbf{p})$ and its corresponding weight $\textbf{w}^t(\textbf{p})$ integrate ${D}_{c}^{t}$ into the TSDF volume. These update functions are truncated before and after the surface to ensure efficient runtime and robust reconstruction of fine-surface details. Further, we mapped surface normal vectors $N_\textrm{ref}^t$ via relative rotation w.r.t the reference frame. Fig.~\ref{fig:overall_pipeline} provides overall system pipeline. 

\RestyleAlgo{ruled}
\begin{algorithm}[t]
\caption{$L_1$ Mobile Robotic MVPS}\label{alg:mobileMVPS}
Set $\{l_{k}^{0}\}_{k=1}^8, \{e_{k}^{0}\}_{k=1}^8 := \textrm{predictLight}(\mathcal{I}_{ps}^{0})$;\\
Set ${N}_\textrm{ps}^{0}, {\Lambda}_\textrm{ps}^{0} := \textrm{predictNormal}(\mathcal{I}_{ps}^{0}, \{l_{k}^{0}\}_{k=1}^8, \{e_{k}^{0}\}_{k=1}^8)$;\\
Set ${D}^{0}_\textrm{pd} := \textrm{predictDepth}(I_\textrm{si}^{0})$;\\
Set ${D}_{c}^{0} := \textrm{refineDepth}({D}^{0}_\textrm{pd}, {N}_\textrm{ps}^{0}, {\Lambda}_\textrm{ps}^{0})$; \textcolor{blue}{/*  Eq.\eqref{eq:overall_optimization_per_view} solution */}\\
Set $\mathbf{P}_\text{ref}^{0} := [\mathbf{R}^{0} | \lambda \hat{\mathbf{t}}^{0}]$;\\
Set $[\mathbf{V}^{0}, \mathbf{W}^{0}] := \textrm{initialize}(D^{0}_{c}, \mathbf{K}, \mathbf{P}_\text{ref}^{0})$;\\
 \For{$t=1:T$}{
    $\{l_{k}^{t}\}_{k=1}^8,  \{e_{k}^{t}\}_{k=1}^8 := \textrm{predictLight}(\mathcal{I}_{ps}^{t})$;\\
    ${N}_\textrm{ps}^{t}, {\Lambda}_\textrm{ps}^{t} := \textrm{predictNormal}(\mathcal{I}_{ps}^{t}, \{l_{k}^{t}\}_{k=1}^8, \{e_{k}^{t}\}_{k=1}^8)$;\\
    ${D}^{t}_\textrm{pd} := \textrm{predictDepth}(I_\textrm{si}^{t})$;\\
    ${D}_{c}^{t} := \textrm{refineDepth}({D}^{t}_\textrm{pd}, {N}_\textrm{ps}^{t}, {\Lambda}_\textrm{ps}^{t})$;\\ 
    $\mathbf{V}^{t-1}_{local} := \textrm{localTSDFExtraction}(\mathbf{V}^{t-1}, \mathbf{W}^{t-1}, D^{t}_{c}, \mathbf{K})$;\\
    $\mathbf{v}^{t}_{local} := \textrm{localTSDFUpdate}(\mathbf{V}^{t-1}_{local}, \mathbf{W}^{t-1}_{local}, D^{t}_{c}, \mathbf{K})$;\\
    $[\mathbf{P}_\text{ref}^{t}, {N}_\text{ref}^{t}, \mathbf{v}^{t}, \mathbf{w}^{t}] := \textrm{poseEst\_globalAlign}(\mathbf{v}^{t}_{local}, \mathbf{V}^{t-1}, \mathbf{W}^{t-1}, {N}_\textrm{c}^{t}, \mathbf{P}_\text{ref}^{t-1})$;\\
     \textcolor{blue}{/* ${N}_\textrm{c}^{t}$ is the normal map corresponding to ${D}_\textrm{c}^{t}$ */}\\
     \textcolor{blue}{/* \citep{yang2020teaser} fast implementation is used for registration*/}\\
    $[\mathbf{V}^{t}, \mathbf{W}^{t}] := \textrm{updateReconstruction}(\mathbf{v}^{t}, \mathbf{w}^{t})$;
 }
\end{algorithm}

\section{Experiments, Results and Limitations}\label{sec:experimentandresults}

Algorithm Tab.~\ref{alg:mobileMVPS} provides pseudo code of our test time setup. The train and test script is coded on pytorch 2.1.2. with few dependencies uses C/C++. \revised{The proposed method is implemented on a computing machine featuring an Apple M4 chip with a 10-core CPU, 10-core GPU, and a 16-core Neural Engine, all running on macOS.}

\smallskip
\formattedparagraph{Train Time Setup.}
We used the light calibration network architecture proposed in \citep{chen2019self}. We trained this network on Blobby and Sculpture datasets \citep{chen2018ps} with Adam optimizer \citep{kingma2014adam} at an initial learning rate of $5 \times 10^{-4}$. The model is trained for 20 epochs with a batch size of 32. The learning rate was divided by two after every 5 epochs. For modeling uncertainty in surface normal prediction model, we use the popular drop-out approach to CNN-PS network \citep{kaya2022uncertainty}. The uncertainty-based normal estimation network is trained using CyclesPS dataset \citep{ikehata2018cnn}, which contains 15 shapes rendered with diffuse, specular, and metallic BRDFs using 1300 light sources. We use 90\% of the CyclesPS dataset for training and 10\% for validation. The observation map of dimension $32 \times 32$ is generated using CyclesPS images to train the normal estimation network. For each observation map, 50 to 1300 light sources are picked randomly. This network is trained for 10 epochs using Adam optimizer \citep{kingma2014adam} with a learning rate of $10^{-1}$. For each pixel, the mean and variance of the outputs are computed to model the surface normal prediction and its uncertainty. For depth prediction, we used the pre-trained MiDaS v3.1 model, which is trained on 12 datasets from MiDaS original setup with a few additional training data as detailed in \citep{birkl2023midas}.

\smallskip
\formattedparagraph{Test Time Setup.}
We used the proposed mobile robotic MVPS hardware setup in an incremental setting at test time. For consistency with the recent state-of-the-art in MVPS \citep{li2020multi, kaya2022uncertainty, kaya2023multi}, we took 5-objects from the DiLiGenT-MV dataset \citep{li2020multi}, namely BEAR, BUDDHA, COW, READING, and POT2, to evaluate our performance. We additionally incorporated a simple SPHERE object to conduct ablations. Our test time MVPS hardware comprises of a camera with a $\text{50~mm}$ lens and 8 LEDs that capture images with $250 \times 250$ resolution. The robot moves to 36 distinct poses, capturing 8 images per view in an incremental setting. A fixed rotation of $10^{\circ}$ is given for every change in the robot pose. The robot navigation path and motion model to perform the 3D acquisition at test time is fixed and is shown in Fig.~\ref{fig:img_acquisition_camera_traj}(b). The robot is constrained to move not more than $\text{12~cm}$ from the object. For experiments, we used the object 2D segmented image mask per view. At test time, the light source information, i.e., $\{l_k^{t}\}_{k=1}^8,  \{e_{k}^{t}\}_{k=1}^8$ is predicted using the trained light calibration network while surface normal $N_\textrm{ps}^{t}$ and associated uncertainty $\Lambda_\textrm{ps}^t$ in surface normal prediction is obtained using trained uncertainty based CNN-PS network \citep{ikehata2018cnn}. Depth map per view at test time is predicted using \citep{birkl2023midas} pre-trained network while refined depth map is obtained by optimizing the Eq.\eqref{eq:overall_optimization_per_view}. To fuse the refined depth map per view in an incremental setting, we use a volumetric grid of size $512^3$ for volumetric fusion. At the end, the marching cubes algorithm \citep{lorensen1987marching} is used to extract the object's 3D mesh for performance evaluation.

\begin{table*}[t]
\small
\centering
    \resizebox{1.0\textwidth}{!}
    {
    \begin{tabular}{c|c|c|c|c|c|c|c|c|c}
   &
   \multicolumn{2}{c|}{\cellcolor{gray!60} Deep Multi-View Stereo (MVS)} &
   \multicolumn{5}{c|}{\cellcolor{gray!60} Deep Multiview Photometric Stereo (MVPS)}&
    \multicolumn{2}{c}{\cellcolor{blue!20} Mobile Robotic MVPS}
    \\ 
    \hline
       \textbf{Dataset}$\downarrow$ &
       \cellcolor{gray!20} MVSNet \citep{yao2018mvsnet} &
       \cellcolor{gray!20} PM-Net \citep{wang2021patchmatchnet} &
       \cellcolor{gray!20} R-MVPS \citep{park2016robust} &
       \cellcolor{gray!20} B-MVPS \citep{li2020multi} &
       \cellcolor{gray!20} NR-MVPS \citep{kaya2021neural} &
       \cellcolor{gray!20} UA-MVPS \citep{kaya2022uncertainty} &
       \cellcolor{gray!20} MVPS-Rev \citep{kaya2023multi} &
       {~~Ours~~} & 
       {~~\textcolor{black}{Ours + IBA \citep{indelman2015incremental}}~~}\\
        \hline
        BEAR     & 0.135 & 0.672 & 0.504  & 0.986 & 0.856 & 0.895  & 0.965  & 0.882 & \textcolor{black}{0.896}\\
        BUDDHA   & 0.147 & 0.799 & 0.935  & 0.934 & 0.690 & 0.922  & 0.993  & 0.911 & \textcolor{black}{0.942}\\
        COW      & 0.095 & 0.734 & 0.915  & 0.989 & 0.844 & 0.979  & 0.987  & 0.914 & \textcolor{black}{0.939}\\
        READING  & 0.115 & 0.834 & 0.869  & 0.975 & 0.720 & 0.970  & 0.975  & 0.897 & \textcolor{black}{0.907}\\
        POT2     & 0.126 & 0.666 & 0.458  & 0.984 & 0.858 & 0.907  & 0.991  & 0.909 & \textcolor{black}{0.914}\\
        \hline
    \end{tabular}
    }
    \caption{3D acquisition result comparison with classical MVS and recent state-of-the-art deep learning-based MVPS approaches on DiLiGenT-MV benchmark dataset \citep{li2020multi}. Despite being an incremental mobile robotic setup, our approach can provide favorable results with just using 8 light varying images per view. In total our approach uses $36~\text{views}~\times8~\text{images per view}= 288$ images compared to other methods that use all $20~\text{views}~\times~96~\text{images per view} = 1920$ images in total to recover 3D shape. We used the F-score metric for the statistical comparison to keep the evaluation metric consistent with the previous methods.
    }
    \label{tab:main_comparison_table}
\end{table*}

\begin{figure*}[t]
\centering
\includegraphics[width=1.0\textwidth]{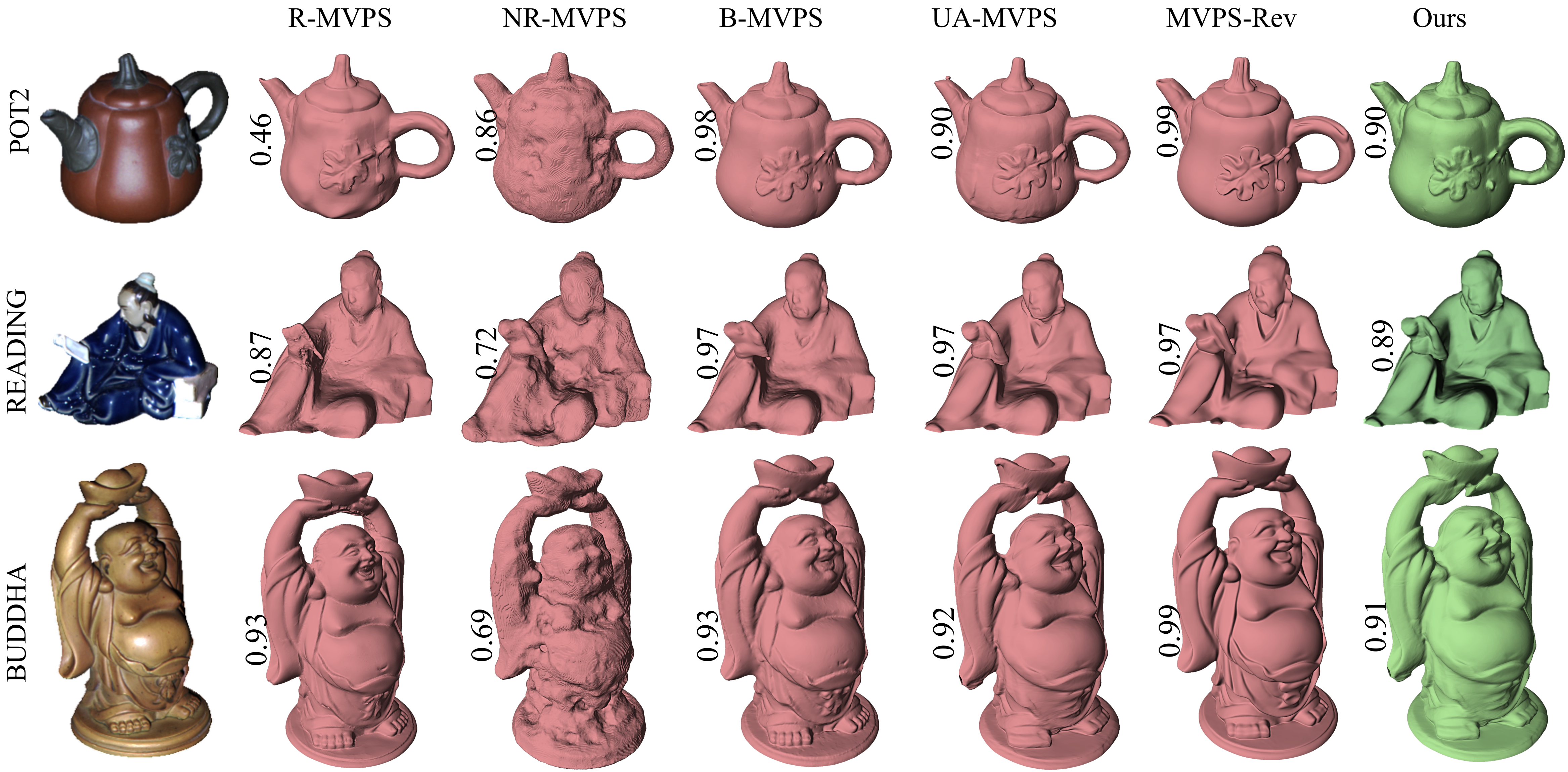}
    \caption{\textbf{Comparison with the state of the art MVPS methods.} We compared our methods to R-MVPS \citep{park2016robust}, NR-MVPS \citep{kaya2021neural}, B-MVPS \citep{li2020multi}, UA-MVPS \citep{kaya2022uncertainty}, and MVPS-Rev \citep{kaya2023multi}.  Despite our setting being different from the existing MVPS, we recover the object's 3D geometry comparable to these methods, showing its suitability for the next step in MVPS, i.e., mobile robotic automation in photogrammetry for fine-detailed 3D acquisition of objects.}
    \label{fig:diligent_comparison_results}
\end{figure*}

\subsection{Statistical Evaluation}
\begin{figure*}[t]
	\centering
	\begin{subfigure}{0.48\linewidth}
		\includegraphics[width=\linewidth]{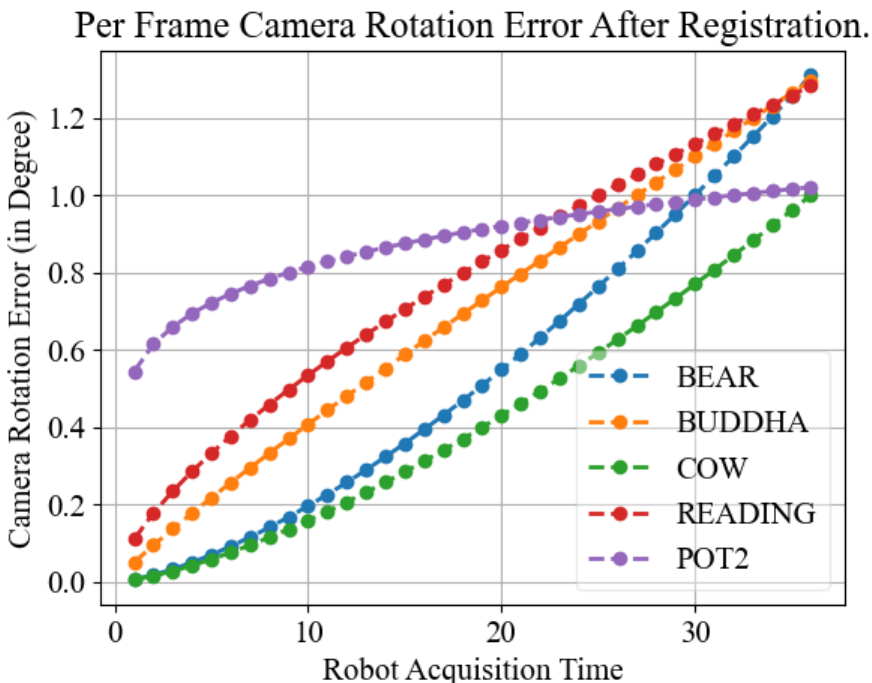}
		\caption{Rotation Error}
		\label{fig:rotation_error}
	\end{subfigure}
	\begin{subfigure}{0.48\linewidth}
		\includegraphics[width=\linewidth]{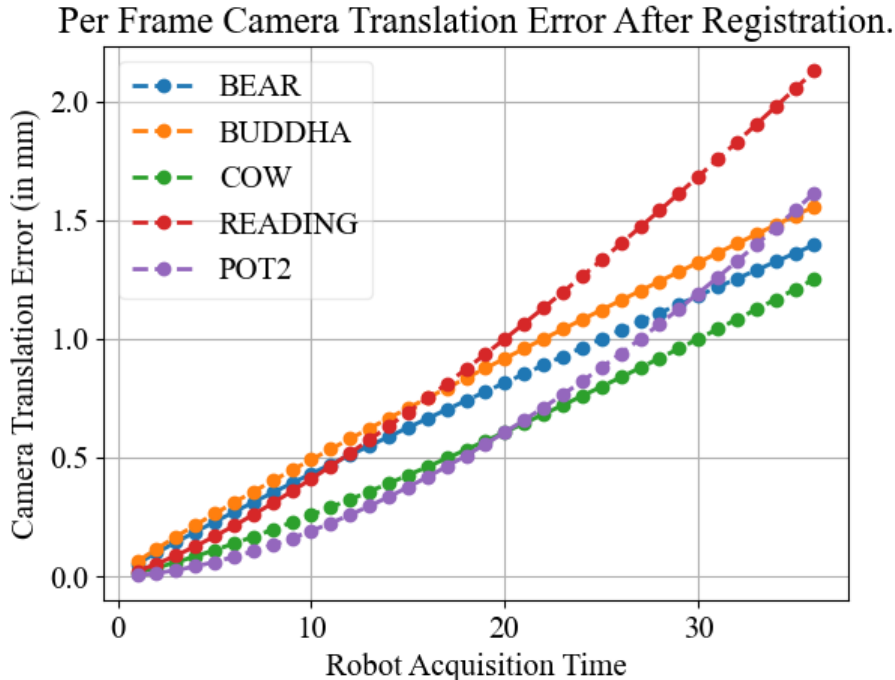}
		\caption{Translation Error}
		\label{fig:translation_error}
	\end{subfigure}
	\caption{\textbf{(a)}-\textbf{(b)} Camera rotation and translation error over frames for each object category after registration, respectively.}\label{fig:approach_evaluation}
\end{figure*}

\smallskip
\formattedparagraph{Evaluation Metric.}
We used the popular F-score metric defined in \citep{knapitsch2017tanks} to evaluate the object's 3D reconstruction accuracy. We used the $L_2$ difference between the ground-truth pose and the estimated pose for camera pose error evaluation after registering the recovered pose to absolute real-world metric values. Note that for consistency with well-known previous approaches, few ablations in this work use average relative depth (ARD) error \citep{lee2019monocular, kumar2019superpixel} and Chamfer $L_1$ \citep{kaya2022uncertainty} to report the results on the benchmark dataset.

\smallskip
\formattedparagraph{\textit{(i)} Comparison with state-of-the-art MVS and MVPS methods.} We compared our approach results with recent deep learning-based multiview stereo (MVS) and multiview photometric stereo (MVPS) approaches. We used the DiLiGenT-MV \citep{li2020multi} dataset for performance comparison. We adopted the F-score metric to quantify our method's performance, facilitating a direct comparison with the state-of-the-art methodologies. The results in Table \ref{tab:main_comparison_table} show our approach's suitability over deep MVS and MVPS techniques, highlighting the effectiveness in more challenging experimental settings, i.e.,  mobile robotic setups. Fig.~\ref{fig:diligent_comparison_results} shows qualitative 3D reconstruction result comparison with state-of-the-art approaches. We can observe that despite mobile setup and using only eight light-varying PS images at a time, we can achieve accuracy that favorably compares to classical MVPS setup-based approaches. Hence, our work provides a constructive future direction for automating the photogrammetry. Moreover, our approach uses $36~\text{views}~\times8~\text{images per view}= 288$ images only compared to other methods, which uses $1920$ images in total. 

\smallskip
\formattedparagraph{\textit{(ii)} Camera pose error over frames at test time.} We compute the camera pose per frame, where each new camera pose is computed by registering the shape to the last frame shape---the previously computed pose is already consistent with the reference, i.e., the first frame. We used \citep{yang2020teaser} to perform the registration due to its impressive computation time and robustness to outliers. Once all 36 camera poses are recovered, we register them to the ground-truth pose trajectory for error computation. Fig.~\ref{fig:rotation_error} and Fig.~\ref{fig:translation_error} show the rotation and translation error observed on the test data, respectively. It can be observed that the computed poses are reasonably good, yet the error accumulates over frames. \revised{To reduce the camera pose accumulation error over frames, we integrated IBA \citep{indelman2015incremental} to the proposed pipeline. This helps in improving the object's 3D reconstruction accuracy (refer Table \ref{tab:main_comparison_table}), increasing the overall computational time nonetheless (refer Table \ref{tab:timingcomparison}). Despite our paper's main focus is to develop an incremental approach, we analyzed a global method, i.e., MVR \citep{pulli1999multiview} to reduce the overall camera pose error for improved 3D reconstruction accuracy. Detailed results and analysis related to use of both IBA \citep{indelman2015incremental} and MVR \citep{pulli1999multiview} method on DiLiGenT-MV dataset \citep{li2020multi} is provided in \textbf{\textit{(ii)}} of \ref{apx:mea}.}

\begin{figure*}[t]
	\centering
	\begin{subfigure}{0.455\linewidth}
	        \includegraphics[width=\linewidth]{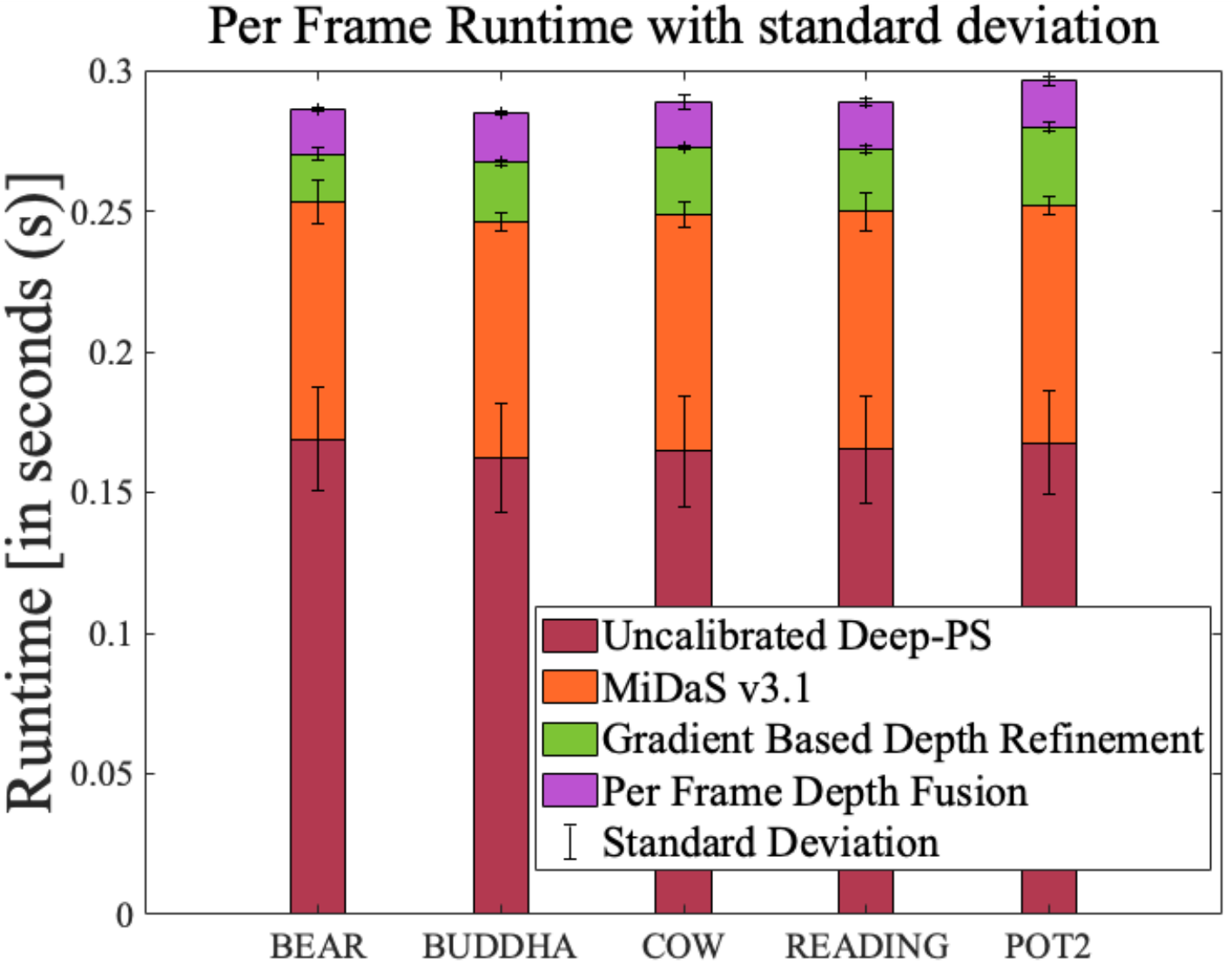}
	        \caption{Runtime}
	        \label{fig:runtime}
    \end{subfigure}
    \begin{subfigure}{0.48\linewidth}
		\includegraphics[width=\linewidth]{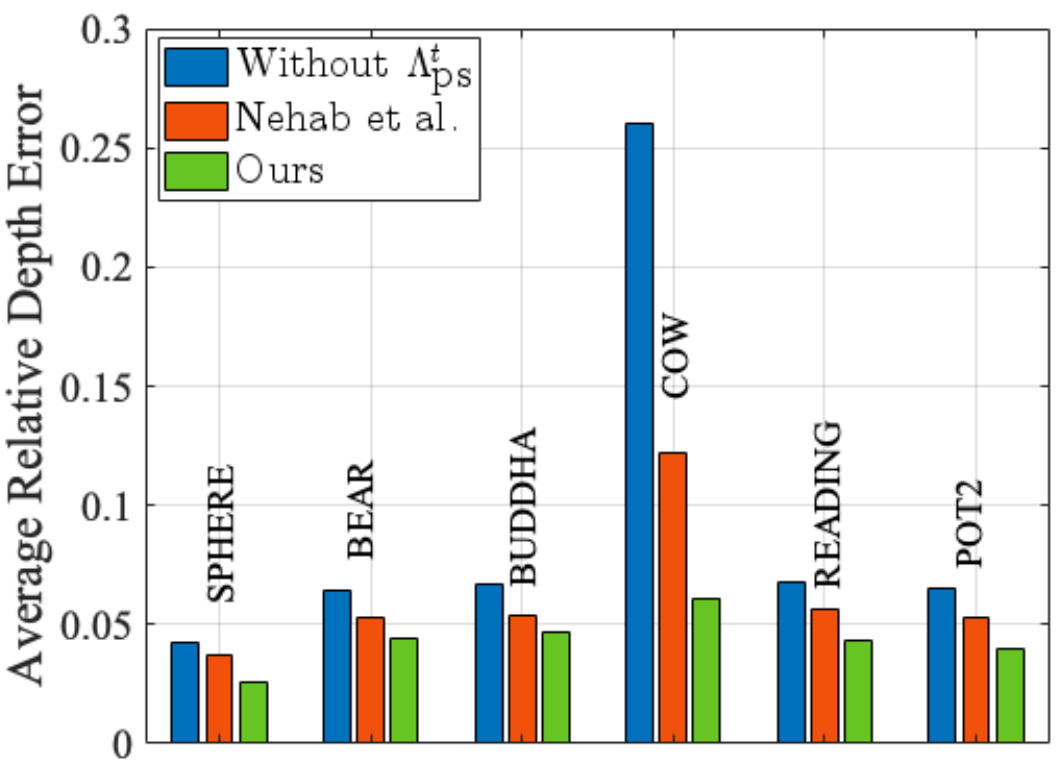}
		\caption{$\Lambda_\textrm{ps}^{t}$ for Depth Refinement}
		\label{fig:nehabcomparison}
	\end{subfigure}     
	\caption{\textbf{(a)} Processing time for one frame at test time. We ran the experiment 300 times to come up with mean and standard deviation in runtime---the standard deviation is shown with a line within the error bar graph. For clarity, we further show the time consumed by each component of our approach with different colors. \textbf{(b)} Effect of data-driven uncertainty modeling in our gradient-based depth map refinement. The average relative depth (ARD) error \citep{lee2019monocular} obtained on different dataset shows our approach effectiveness. }\label{fig:approach_evaluation1}
\end{figure*}

\smallskip
\formattedparagraph{\textit{(iii)} Computational time.}  We noted the overall computational time to study the suitability of our approach for mobile robotic applications. Fig.~\ref{fig:runtime} shows per-frame processing time consumed by each independent component proposed in the paper on respective object categories. We also computed the possible variations in computational time by running the same experiments 300 times and documenting the standard deviation (refer Fig.~\ref{fig:runtime}). Quantitatively, we can fuse our refined depth per frame to the global volume in approximately—0.3 seconds (s), thus making it suitable for robotics applications. \revised{Meanwhile, we compared our method's processing time with recent state-of-the-art MVPS methods, demonstrating its efficiency. For this experimental comparison, we provide same number of images as input to all the methods. Table \ref{tab:timingcomparison} provides the statistical timing comparison results (in seconds), clearly showing the applicability of our approach for robotics application. Note, however, that we added IBA \citep{indelman2015incremental} to our pipeline to reduce the accumulation of error over frames, which leads to an increase in the overall computational time, improving the overall 3D reconstruction accuracy nonetheless---cf. Table \ref{tab:main_comparison_table}. Our efficiency in computational time stems from processing fewer frames, use of single image depth priors, and introduction of faster optimization approach for a mindful fusion.}

\begin{figure*}[t]
	\centering
	\begin{subfigure}{0.48\linewidth}
		\includegraphics[width=\linewidth]{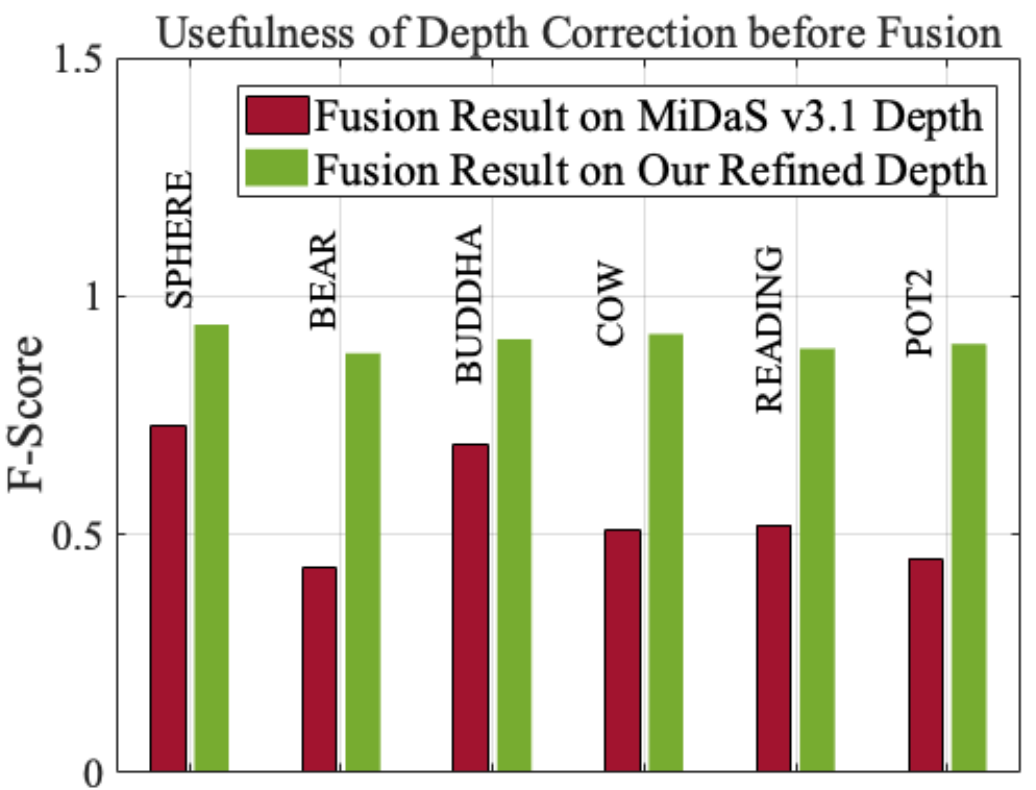}
		\caption{Refined Depth Fusion}
		\label{fig:depthfusionrefinement}
	\end{subfigure}
	\begin{subfigure}{0.48\linewidth}
	        \includegraphics[width=\linewidth]{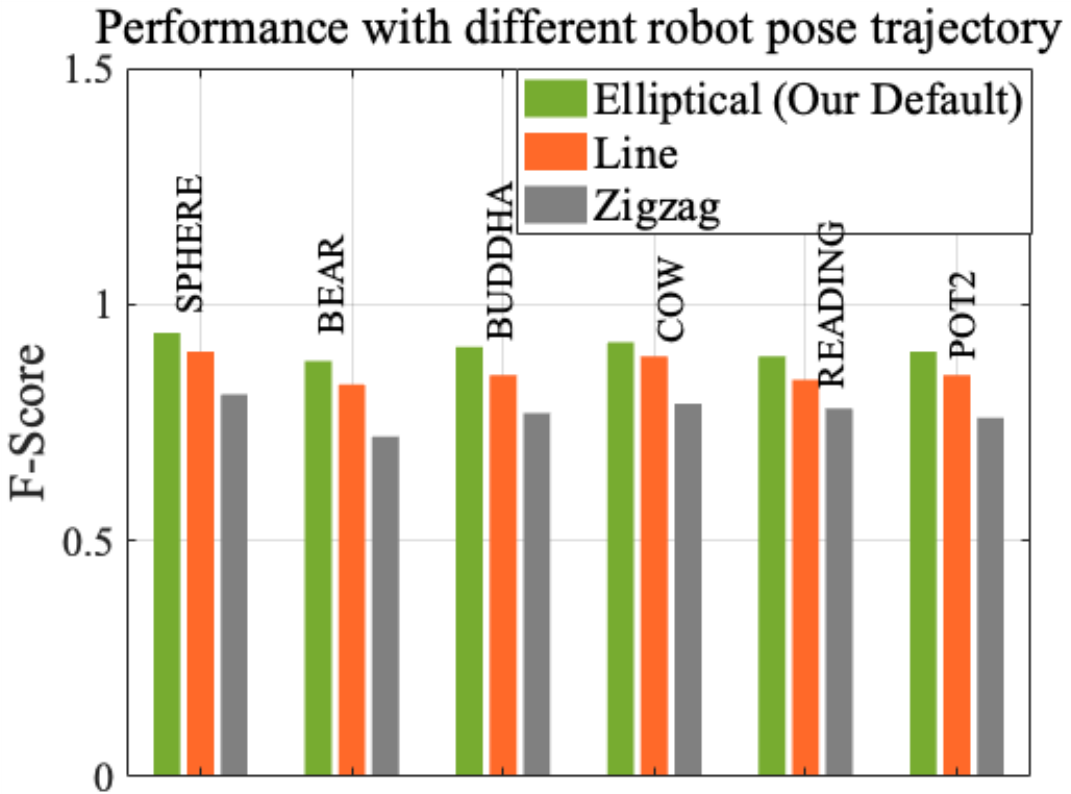}
	        \caption{Robot Trajectory Type}
	        \label{fig:robottrajectoryresult}
         \end{subfigure}
	\caption{\textbf{(a)} Benefit of correcting the low-frequency and high-frequency bias due to the PS surface normal and global depth map, respectively. Overall, the F-score clearly shows the benefit in overall performance. \textbf{(b)} Result using different robot camera pose trajectory. Here, we show a couple of trajectory examples and their respective results.}
	\label{fig:approach_evaluation2}
\end{figure*}

\begin{table}[h]
    \centering
    \resizebox{1.0\textwidth}{!}
    {
    \textcolor{black}{
		\begin{tabular}{c|c|c|c|c|c}
			\hline
			 \textbf{Dataset} $\downarrow$ & NR-MVPS \citep{kaya2021neural} & UA-MVPS \citep{kaya2022uncertainty}  & MVPS-Rev \citep{kaya2023multi}  & Ours & Ours + IBA \citep{indelman2015incremental}  \\ \hline
			  BEAR & 18001.23 & 305.25 & 365.62 & 0.2862  & 141.62 \\ \hline
			BUDDHA & 19440.56  & 302.19 & 359.51 & 0.2851 & 140.21 \\ \hline
            COW & 21348.40  & 316.73 & 384.27 & 0.2887 & 142.63 \\ \hline
            READING & 22032.97  & 328.28 & 402.28 & 0.2889 & 142.77 \\ \hline
            POT2 & 19908.33  & 309.43 & 373.42 & 0.2962 & 142.94 \\ \hline
		\end{tabular}
        }
    }
	\caption{\textcolor{black}{Processing time comparison with recent state-of-the-art MVPS methods on DiLiGenT dataset in our experimental setting. Note that we provided same number of images to all the methods for the experimental comparison. All the timings listed above are in seconds(s).}}\label{tab:timingcomparison}
\end{table}

\subsection{Ablation Study}

\smallskip
\noindent
\formattedparagraph{\textit{(i)} Usefulness of Uncertainty Modeling.} \citep{nehab2005efficiently} is a well-known work that used depth prior to improve surface normal and surface normal prior to improve overall depth.  Yet, it uses pre-defined convolutional filters to refine the depth solution via surface normal prior. On the contrary, we use a data-driven approach to learn the uncertainty in surface normal prediction at test time and refine the overall depth---refer to Eq.\eqref{eq:overall_optimization_per_view}. We verify the effectiveness of our depth refinement optimization by comparing our depth refinement result with experimental results obtained under the following two settings: \textbf{(I)} Refine depth using Eq.\eqref{eq:overall_optimization_per_view} without the predicted uncertainty variable, i.e., $\Lambda_\textrm{ps}^{t}$. \textbf{(II)} Use Nehab et al. \citep{nehab2005efficiently} approach to refine overall depth per view. Fig.~\ref{fig:nehabcomparison} shows the quantitative result comparison on different datasets, showing our approach's effectiveness. The result clearly shows that the design choice used in \citep{nehab2005efficiently}, i.e., the use of surface normal smoothing and handcrafted convolutional kernel to refine depth, may not generalize well across different experimental settings.

Moreover, we conducted an experiment where we fused the MiDaS v3.1 depth at test time per frame to recover the object's 3D geometry. We compared the recovered object geometry to the one we obtained using our approach. Fig.~\ref{fig:depthfusionrefinement} show the F-score of the recovered objects, thereby further demonstrating the usefulness of uncertainty modeling in Eq.\eqref{eq:overall_optimization_per_view} optimization for mindful depth fusion.

\smallskip
\formattedparagraph{\textit{(ii)} Effect of robot's pose trajectory at test time.} In addition to our default navigation path (see Fig.~\ref{fig:img_acquisition_camera_traj}(b)), we conducted a couple of more experiments, where the robot is allowed to move in varied camera trajectory path such as a line and in a zigzag path, yet the camera is configured to face the object for capturing the object's multiple viewpoints (refer supp. for camera path details). Note that the robot uses a pre-defined motion model. We computed the F-score of the recovered 3D geometry and compared it to our results obtained using the default robot trajectory path. Fig.~\ref{fig:robottrajectoryresult} shows the results obtained under different camera pose trajectory. Our observation from this experiment is that although our default setting gives the best result, the difference in results are not great, demonstrating our approach's robustness. We observed that the 3D reconstruction error increases mainly due to the camera pose estimation error accumulation over frames for different kinds of robot movement. Note that there could be many robot cameras pose trajectories for which we may get better results. Yet, to come up with the best possible camera trajectory is beyond paper's main focus.

\begin{table}[h]
    \centering
    \resizebox{\textwidth}{!}
    {
		\begin{tabular}{c|c|c|c|c|c}
			\hline
			\textbf{Setup}$\downarrow$ $|$ \textbf{Dataset} $\rightarrow$ & ~BEAR~ & ~BUDDHA~ & ~COW~  & ~READING~&   ~POT2~ \\ \hline
			  Predicted Light & 0.418 & 0.542 &	0.427 &	0.419 &	 0.612\\ \hline
			Ground-truth light &  0.394 & 0.562 & 0.433 &	0.428 &	 0.597 \\ \hline
		\end{tabular}
    }
	\caption{Inferred light at test time performs equally well, and the difference in results is not significant. The above statistics show Chamfer $L_1$ distance metric (Lower is better)}\label{tab:light_calibration_table}
\end{table}

\smallskip
\formattedparagraph{\textit{(iii)} Effect of Light Calibration Parameters.} We performed this experiment to validate the reliability of the neural network light calibration model. Note, however, that previous approaches used one viewpoint and allowed table-top setup to rotate the object. Hence, light calibration used to be highly reliable in those MVPS settings. In our setup, the acquisition setup is moving; therefore, this experiment is crucial from a practical standpoint. Table~\ref{tab:light_calibration_table} provides the object's 3D acquisition Chamfer $L_1$ distance result comparison when the predicted light and ground-truth light data are used, respectively. Remarkably, the light direction and intensity predicted by the deep-network provides results that are robust to camera movement, demonstrating its suitability to online MVPS approach for robotic automation.

\smallskip
\formattedparagraph{\textit{(iv)} Robustness to uncontrolled illumination.} 
We studied the robustness of our system performance w.r.t change in the number of external light in the scene. Table \ref{tab:lightvs} provide F-score results showing the performance change with increased external light sources, simulating a good case for uncontrolled illumination. It is quite clear from the 3D reconstruction results that increase in the number of external light source greatly affects the overall reconstruction results.

\begin{table}[h]
    \centering
    {
		\begin{tabular}{c|c|c|c|c}
			\hline
			\textbf{Data}$\downarrow$ $|$ \textbf{No. of Ext. light source} $\rightarrow$ & 0-Ext & 1-Ext & 2-Ext  & 3-Ext   \\ \hline
			  BEAR & 0.882 & 0.784 & 0.522 &  0.441\\ \hline
			BUDDHA & 0.911  & 0.882 & 0.721 & 0.539 \\ \hline
		\end{tabular}
    }
	\caption{Performance (F-score) with increase in number of external (Ext) light source, simulating uncontrolled illumination.}\label{tab:lightvs}
\end{table}

\smallskip
\formattedparagraph{\textit{(v)} Performance variation w.r.t number of LED light source used per view.}
Theoretically, 3 light source should be sufficient ($\mathbf{I} = \mathbf{N}^{T}\mathbf{L}$, R. J. Woodham 1980 \citep{woodham1980photometric}) for our setup. Yet, we used more lights in our robotic sets as 6-8 lights give much better results in practice. Adding more light can be beneficial but the hardware spacing is insufficient to accommodate more than 8 lights in our current robotic setup. Table \ref{tab:ledvs} provides F-score variation w.r.t the number of LEDs fired mindfully, i.e, avoid firing the neighboring LEDs for \#LEDs = 3 and 4. 

\begin{table}[h]
    \centering
    {
		\begin{tabular}{c|c|c|c|c|c|c}
			\hline
			 \textbf{No. of LEDs} $\rightarrow$ & 3 & 4 & 5  & 6 & 7 & 8   \\ \hline
			  BEAR & 0.491 & 0.506 & 0.767 & 0.803  & 0.868 & 0.882\\ \hline
			BUDDHA & 0.485  & 0.667 & 0.806 & 0.890 & 0.902 & 0.911\\ \hline
		\end{tabular}
    }
	\caption{Performance (F-score) with number of LEDs fired.}\label{tab:ledvs}
\end{table}

\subsection{Limitations and Future Direction}
Our approach assumes that the object is placed in a room with a limited lighting condition. In general, with indoor room lighting conditions and multiple light sources, the proposed system may provide inferior 3D acquisition results. Moreover, for this work, we assume the robot's pre-defined motion model and navigation path, and thus, a further interesting direction is to solve the current problem in a fully autonomous setting. Meanwhile, our work \revised{is designed to handle single small and mid-range object size, where, pose-driven} object's 2D image mask is given and, so, benefiting from recent deep learning models such as \citep{kirillov2023segany} \revised{and extending our work for multiple as well as large size objects} in a mobile robotic MVPS setup will surely be helpful for fully autonomous MVPS based photogrammetry system.

\section{Conclusion}
With the recent development in reliable uncalibrated deep-learning approaches to surface normal prediction and single image depth prediction, an effort to automate MVPS for 3D acquisition looks like a natural next constructive step in photogrammetry. To this end, we proposed a reliable mobile robotic system for high-quality 3D acquisition of an object based on the MVPS working principle. This further allowed us to endeavor an incremental approach, a marked departure from the conventional global methodologies previously dominant in the field. The promising results of the proposed approach signify a useful progression in photometric methods for 3D data acquisition, particularly with the goal of automation, where the acquisition setup can have unconstrained movement to perform 3D recovery. The blueprint presented in this work will open numerous avenues for further research in 3D acquisition automation.

\section*{Acknowledgment}
\noindent
Firstly, the author greatly thanks Carlos Oliveira (ETH Zürich) for helping with the robotic system design and for taking the time to render the radiometric images, arrange the images in the correct order, and provide the lighting information in the proper order. Very special mention must be made of MS thesis candidate Noah Zegna Rothenberger (ETH Zürich), who made an initial effort on this project titled ``Bringing Multi-View Photometric Stereo and Mobile Robotics Together'' \citep{noah22bringing}, which provided invaluable insights into the challenges with camera pose recovery and noisy 3D priors inferred from deep learning models. For sharing \citep{kaya2021uncalibrated} code, the author thanks Dr. Berk Kaya (ETH Z\"urich). Lastly, the author gratefully acknowledges the support from Prof. Dr. Luc van Gool (em. KU Leuven, em. ETHZ, INSAIT Sofia University).

\clearpage

\appendix
\section{Abstract}
\paragraph{Following our main draft, in the appendix we provide explicit details of our mobile robotic MVPS hardware here. Next, additional experimental results and ablations on the DiLiGenT benchmark dataset \citep{li2020multi} using our setup are provided, demonstrating the suitability of our mobile robotic MVPS system. Lastly, a brief discussion on the choice of fusion techniques used in the paper and additional benefits of our proposed system is provided for completion, as well as possible future direction for extension}

\section{MVPS Hardware Details}
\begin{figure*}[h]
    \centering
\includegraphics[width=1.0\textwidth]{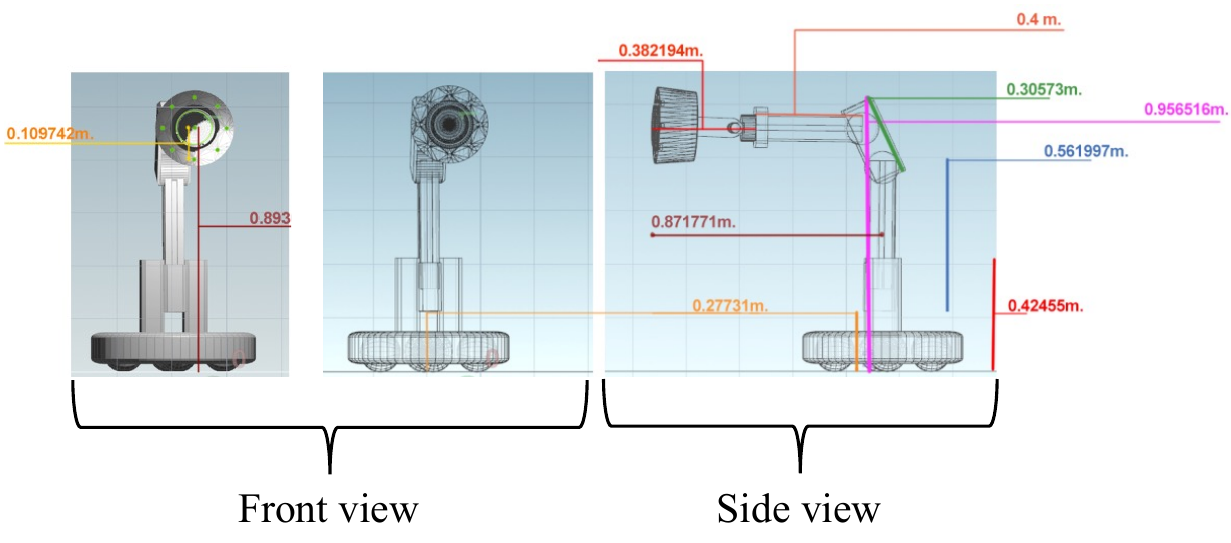}
    \caption{Front and side view of our mobile robotic MVPS setup. Explicit dimension of our design parts shown via CAD model.}
    \label{fig:hardware_mvps}
\end{figure*}

Within the hardware details disclosure limit, we provide a few design details in Fig.~\ref{fig:hardware_mvps} that we used for the 3D data acquisition at test time, demonstrating our design's suitability and portability in automating the MVPS principles in detailed 3D acquisition.

\begin{figure*}[t]
	\centering
	\begin{subfigure}{0.5\linewidth}
		\includegraphics[width=\linewidth]{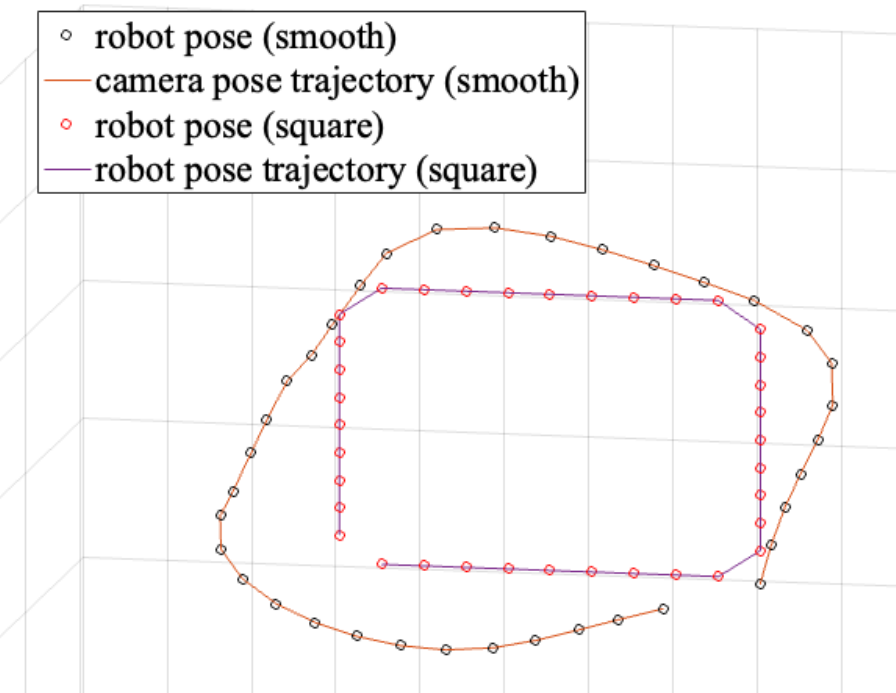}
		\caption{Smooth and Square Trajectory}
		\label{fig:smooth}
	\end{subfigure}
	\begin{subfigure}{0.42\linewidth}
		\includegraphics[width=\linewidth]{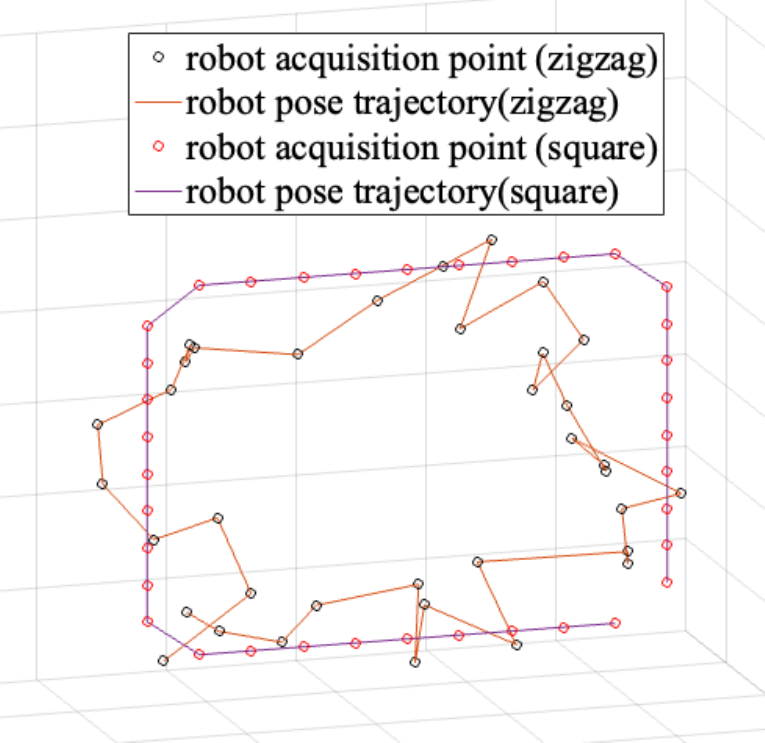}
		\caption{Zigzag and Square Trajectory}
		\label{fig:zigzag}
	\end{subfigure}
	\caption{\textbf{Top view}: Mobile robotic MVPS trajectory at test time. The small circle shows the position where the robot stops and capture the 8 PS images. In addition to the default trajectory denoted as smooth, we additionally test our approach on other robot pose trajectory, namely square and zigzag. We noted that the accuracy of the acquired 3D geometry between distinct trajectory types can be different, yet the difference were not significant as shown in the main paper. }\label{fig:different_robot_trajectory}
\end{figure*}

The data acquisition setup employed at the test time is shown---utilizing a mobile robotic arm, in Fig.~\ref{fig:hardware_mvps}. Continuing from the main paper, we restate that the terminal part of this robotic arm is equipped with a circular arrangement of 8 LED lights equidistantly positioned around a monocular camera, whereby each light maintains an identical distance from the camera's center. This camera is distinguished by a lens possessing a 50mm focal length. Furthermore, the robotic arm's base is mounted on wheels, facilitating the autonomous movement of the robotic assembly. Our approach contrasts with all the previous MVPS benchmark that uses a classical MVPS setup, i.e.,  a stationary camera with LEDs setup and objects positioned on an automated turntable, for example DiLiGenT-MV \citep{li2020multi}. Our approach captures images at test time from varying camera viewpoints, at a different distance from the object and orientations, while the subject within the scene remains immobile. During the test phase, the mobile robot navigates around the object, adopting diverse camera pose trajectories and stops at 36 distinct poses---refer Fig.~\ref{fig:smooth} for a couple of example camera trajectories. A $10^\circ$ separation demarcates the intervals of change in camera pose between these stop points. 8 PS images are captured at each stop point\footnote{stop points are shown with unfilled circle in Fig.~\ref{fig:different_robot_trajectory}}, resulting in a total of 288 images per object (36 poses $\times$ 8 images per pose).

Fig.~\ref{fig:hardware_mvps} further shows the front and side view of the hardware design part, concretely outlining the dimension of the mobile robot used for experimentation.

\section{More Experimental Analysis}\label{apx:mea}
\begin{figure*}
    \centering
\includegraphics[width=\textwidth]{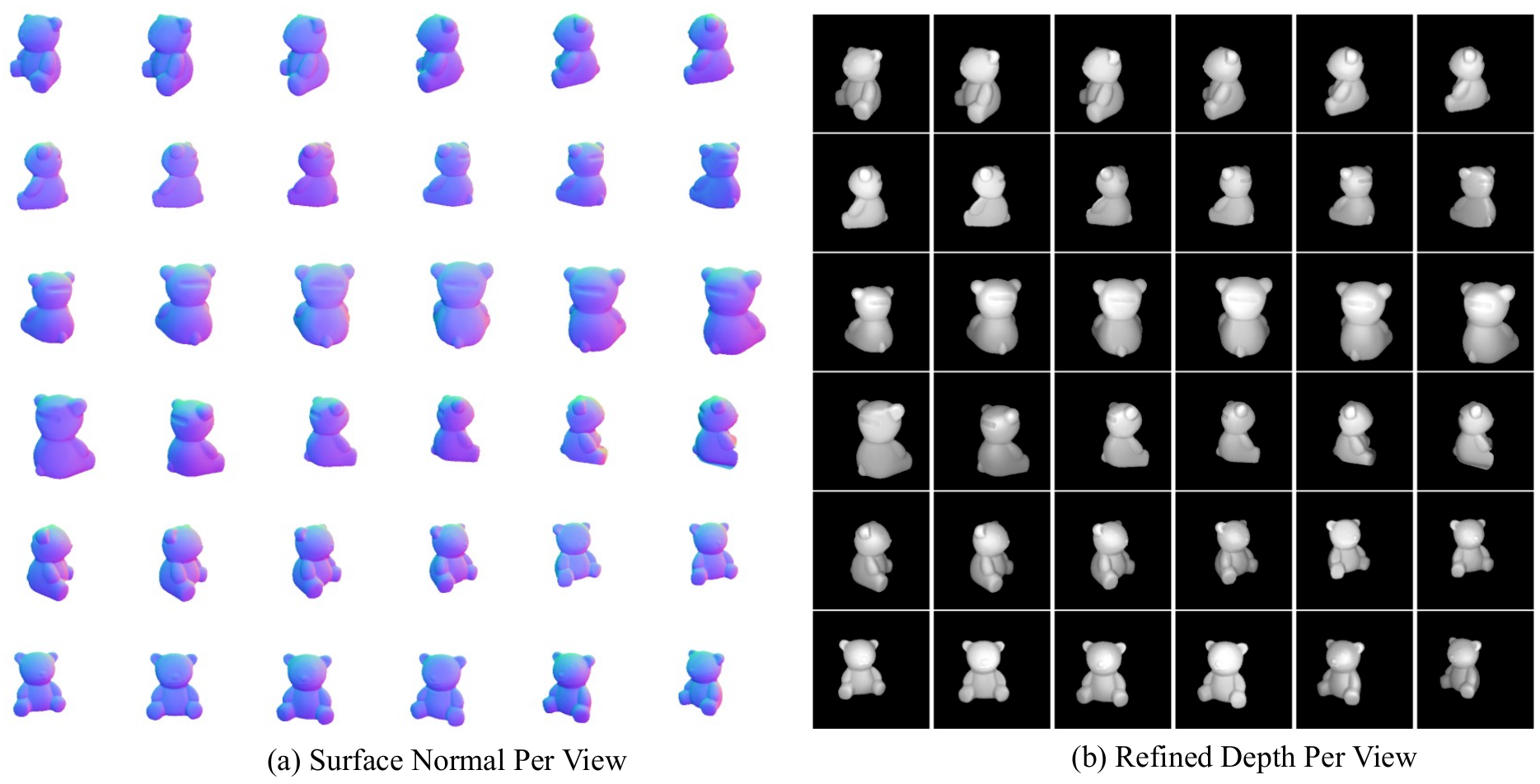}
    \caption{ \textbf{Left}: The object’s surface normal predicted for the all the image taken from 36 different view-points. \textbf{Right}: respective pose depth map per view recovered after Eq.(3) optimization introduced in the main paper.
    }
    \label{fig:refined_depth_results_bear}
\end{figure*}

\formattedparagraph{\textit{(i)} More on the effects of robot pose trajectory.} If we move the robot in a perfect circle and post-process all the images rather than incrementally reconstructing the object, our mobile robotic setup can be considered equivalent to a classic MVPS setup \citep{hernandez2008multiview}. To put it upfront, our prime goal is to overcome such restrictions with MVPS setup and still be able to avail the benefits of MVPS in 3D data acquisition from images using a robotic setup. Furthermore, we want our approach to work for other types of robot trajectories. And therefore, in addition to the default camera pose trajectory shown in the main paper for test time performance evaluation, we further tested our approach on a few other camera pose trajectories. Fig.~\ref{fig:zigzag} shows a couple of camera pose trajectories named `square' and  `zigzag.' As the name `square' suggests, the robot maneuvers a square trajectory at test time in this setting. In comparison, the zigzag trajectory setting allows the robot to move back and forth towards and away from the object. This makes `zigzag' maneuvers particularly challenging, where the successive image could encounter a significant object scale change between two stop points while capturing PS images (see Fig.~\ref{fig:refined_depth_results_bear} as well as supplementary video visual results on scale changes). Note that the accuracy in 3D data acquisition does change with the different trajectory types, yet the difference is not very significant---refer Fig.~\textcolor{red}{6b} in the main paper---showing our approach's applicability.

\begin{figure*}[t]
	\centering
	\begin{subfigure}{0.5\linewidth}
		\includegraphics[width=\linewidth]{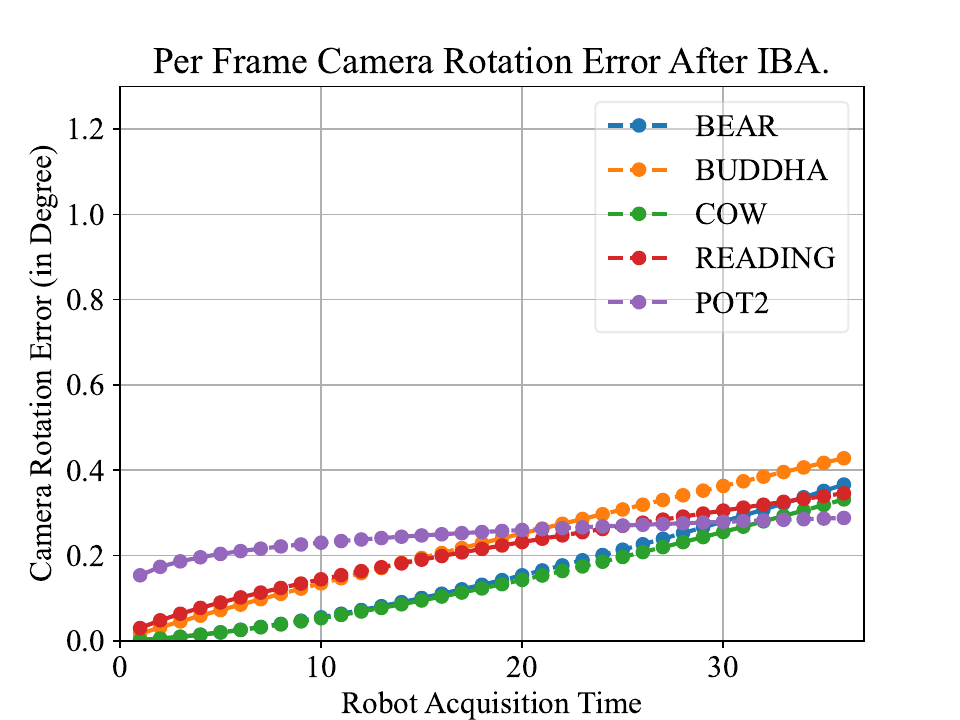}
		\caption{Rotation Error after introducing IBA \citep{indelman2015incremental}}
		\label{fig:rotation_error_IBA}
	\end{subfigure}
	\begin{subfigure}{0.49\linewidth}
		\includegraphics[width=\linewidth]{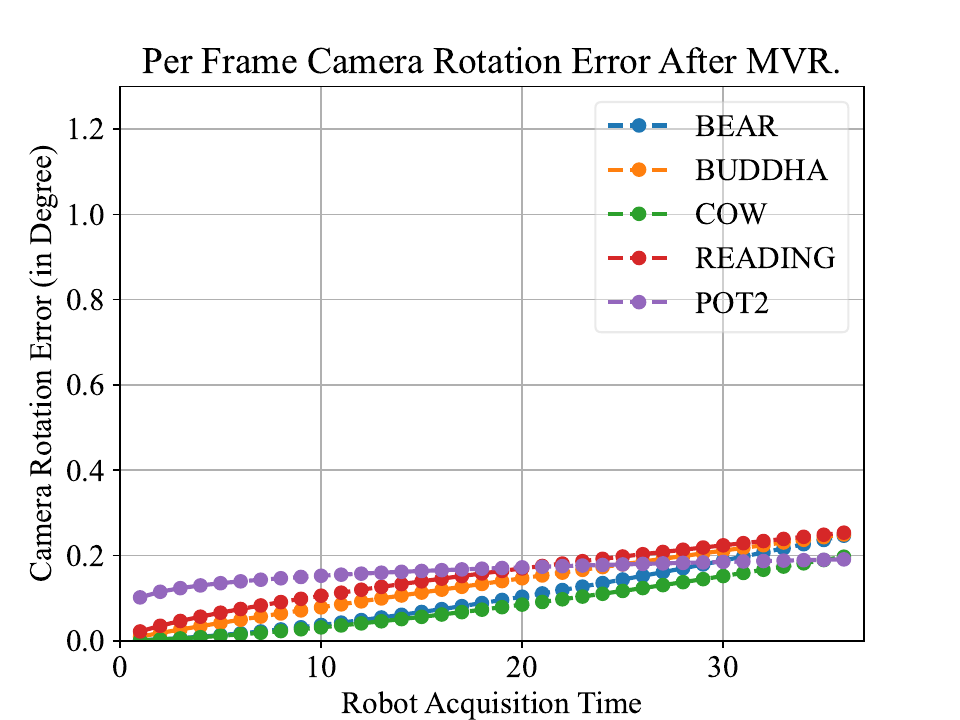}
		\caption{Rotation Error after introducing MVR \citep{pulli1999multiview}}
		\label{fig:rotation_error_MVR}
	\end{subfigure}
	\caption{Effect of using incremental lightweight bundle adjustment (IBA) \citep{indelman2015incremental} and Multi-view Registration (MVR) \citep{pulli1999multiview} method in reducing the overall camera rotation error. Here, IBA focuses on an incremental approach whereas MVR takes a global approach to reduce the overall camera rotation error.  }\label{fig:camera_rotation_error_reduction}
\end{figure*}

\begin{figure*}
	\centering
	\begin{subfigure}{0.49\linewidth}
		\includegraphics[width=\linewidth]{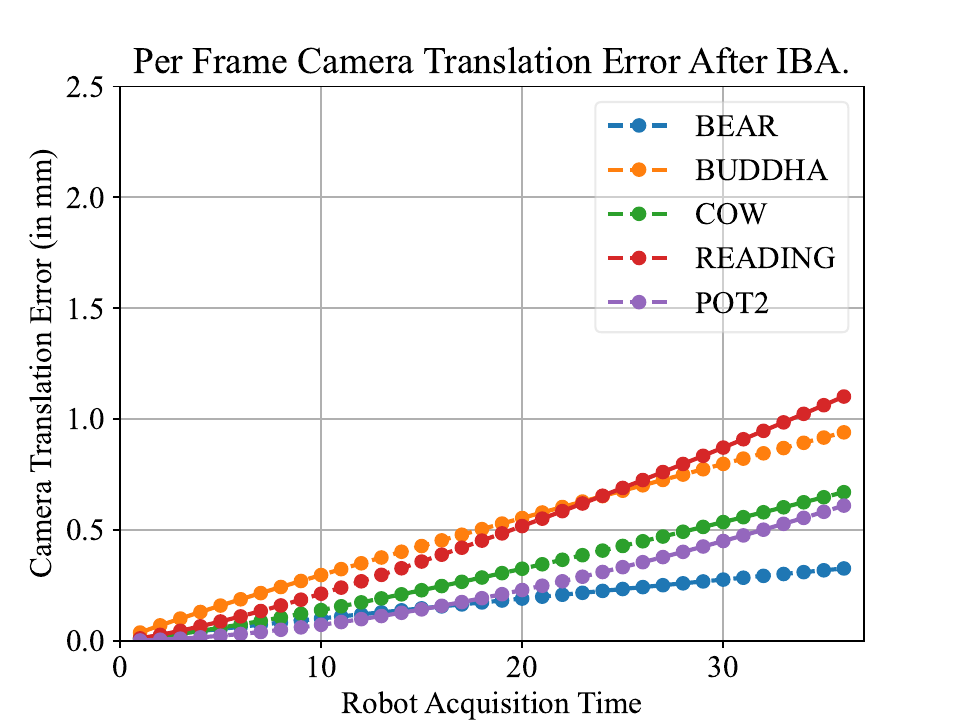}
		\caption{Translation Error after introducing IBA \citep{indelman2015incremental}}
		\label{fig:translation_error_IBA}
	\end{subfigure}
	\begin{subfigure}{0.49\linewidth}
		\includegraphics[width=\linewidth]{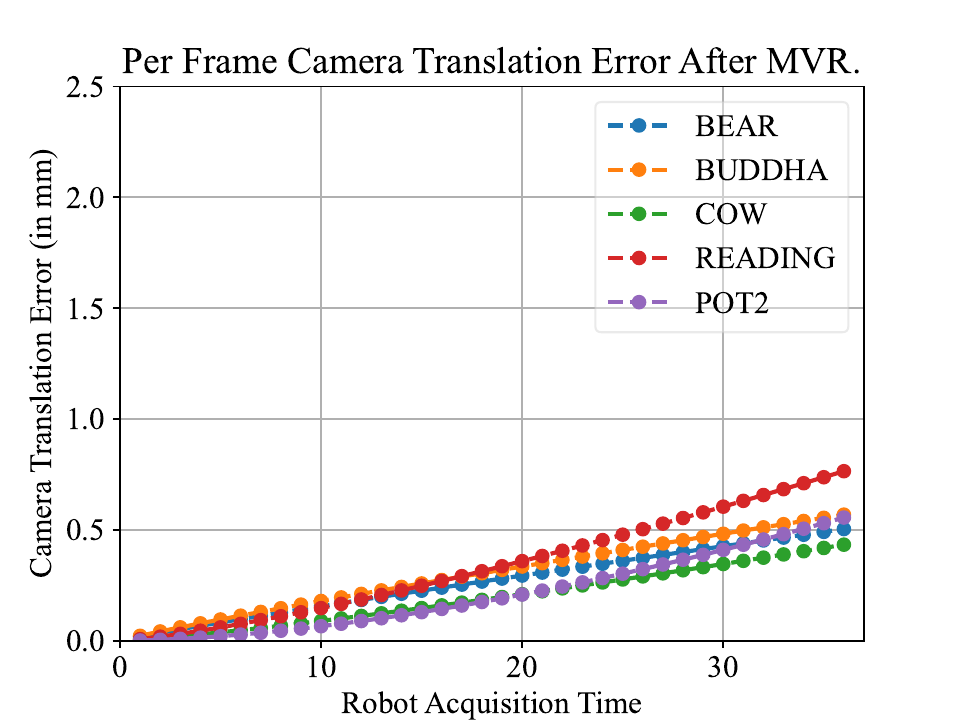}
		\caption{Translation Error after introducing MVR \citep{pulli1999multiview}}
		\label{fig:translation_error_MVR}
	\end{subfigure}
	\caption{Effect of using incremental lightweight bundle adjustment (IBA) \citep{indelman2015incremental} and Multi-view Registration (MVR) \citep{pulli1999multiview} method in reducing the overall camera translation error. Here, IBA focuses on an incremental approach whereas MVR takes a global approach to reduce the overall camera translation error. }\label{fig:camera_translation_error_reduction}
\end{figure*}

\begin{figure*}[t]
    \centering
\includegraphics[width=\textwidth]{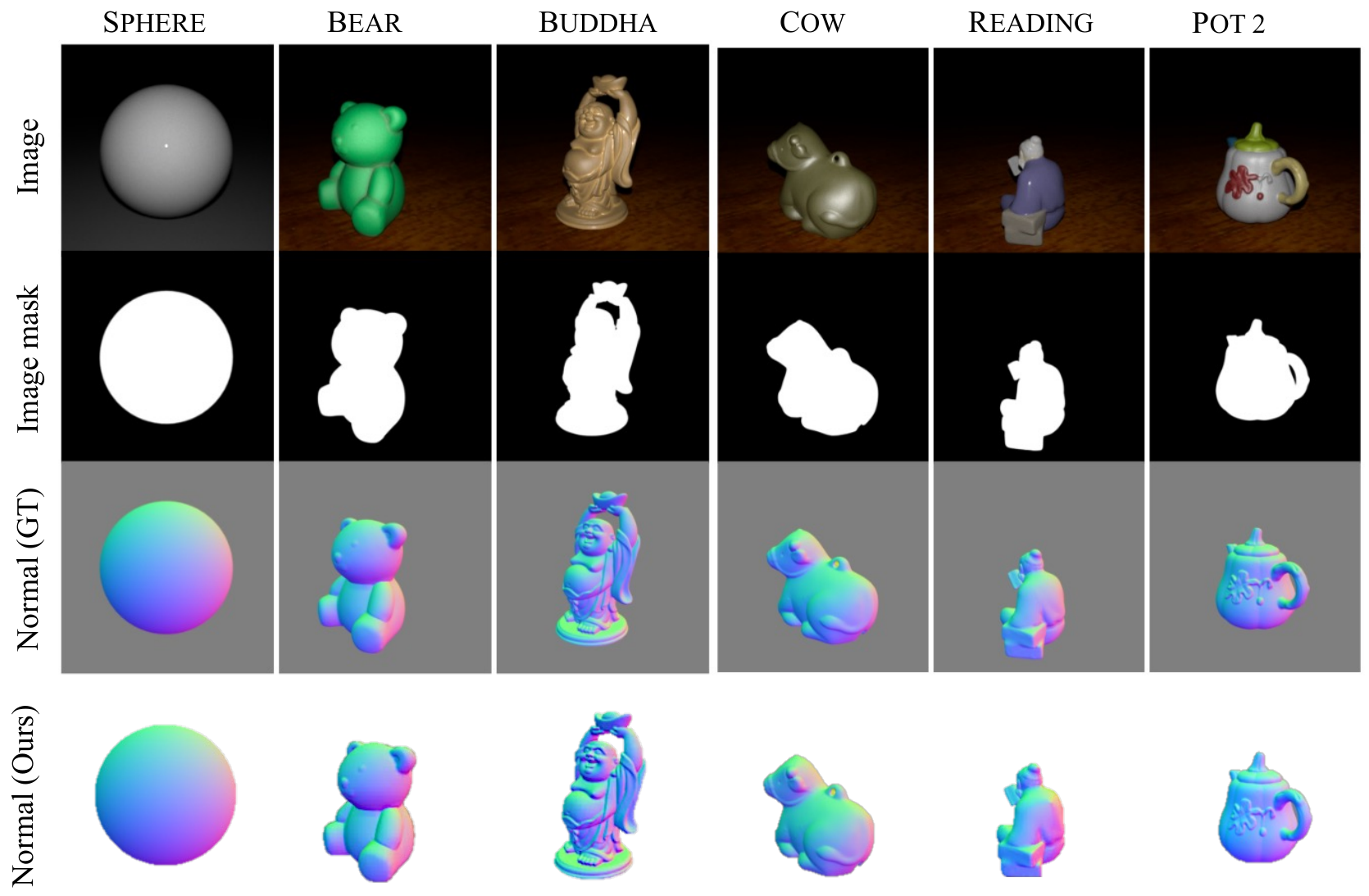}
    \caption{All DiLiGenT benchmark objects \citep{li2020multi} that were tested using our robotic setup. \textbf{Top to Bottom}: one image per object captured for each object category followed by ground-truth 2D image segmentation mask, ground-truth surface normal, and surface normal recovered at test time, respectively.}
    \label{fig:normal_results_one_view}
\end{figure*}

\smallskip
\noindent
\revised{\formattedparagraph{\textit{(ii)} Handling Camera Pose Accumulation Error over Frames.} Despite global methods to improve the camera pose error leading to an offline approach (post-processing backend module), we investigated both incremental as well as global methods for this task. For the incremental method, we explored IBA \citep{indelman2015incremental}, whereas for the global method, we implemented MVR \citep{pulli1999multiview}. MVR leverages pairwise alignments between overlapping views as constraints in the multiview registration process. These constraints effectively distribute registration errors across all views, improving overall alignment accuracy without requiring all data to be loaded simultaneously. For the results presented in Figure \ref{fig:rotation_error_MVR} and Figure \ref{fig:translation_error_MVR}, it is clear that MVA significantly improves the overall performance accuracy. Yet, it leads to an offline approach nonetheless. Consequently, we integrated IBA \citep{indelman2015incremental} to our pipeline to reduce the pose accumulation error over frames for our online approach. IBA is particularly well-suited for our scenarios with sequential image capture or incremental data availability, as it reduces latency demands without sacrificing much to camera pose accuracy compared to MVA (refer Figure \ref{fig:rotation_error_IBA} and Figure \ref{fig:translation_error_IBA}).}

\smallskip
\noindent
\formattedparagraph{\textit{(iii)} Additional results.} Fig.~\ref{fig:normal_results_one_view} show qualitative surface normal result in comparison to the ground-truth surface normal on DiLiGenT-MV dataset object obtained using our setup. The result shows a single view point surface normal results on all the object categories at test time. 

\begin{figure*}
    \centering
\includegraphics[width=\textwidth]{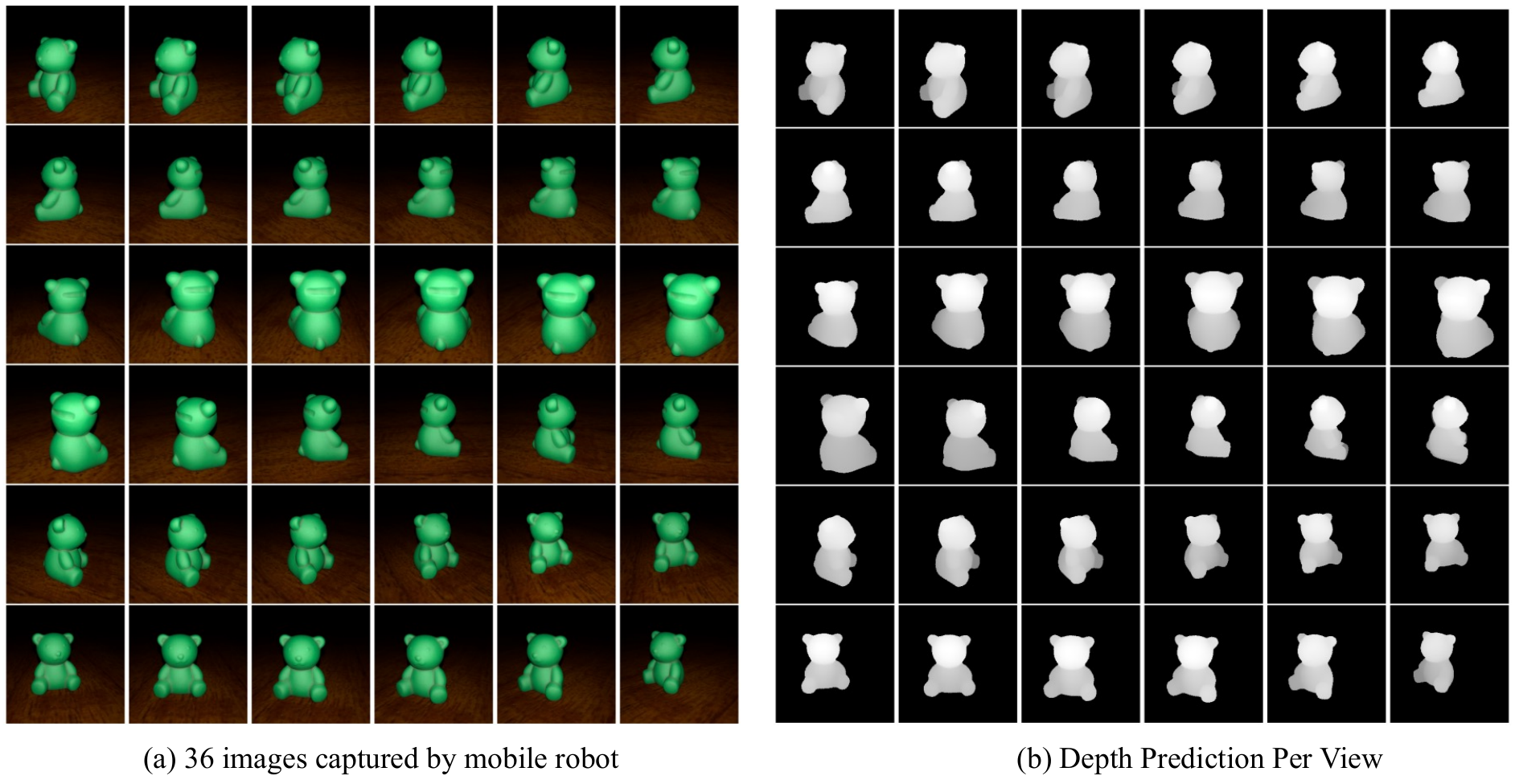}
    \caption{Our mobile robot captures images at a varying distance from the object. \textbf{Left}: showing one of the PS images per view captured at the test time for the BEAR object using our robotic setup. \textbf{Right}: Inferred depth map of the object for the respective views shown on the left \citep{birkl2023midas}.}
    \label{fig:depth_results_bear}
\end{figure*}

Fig.~\ref{fig:depth_results_bear} and Fig.~\ref{fig:refined_depth_results_bear} show test time per view results. Fig.~\ref{fig:depth_results_bear} (\textbf{Right})~show the per-view depth map inferred at test time,  followed by Fig.~\ref{fig:refined_depth_results_bear} that shows the per-view object surface normal and refined depth map obtained after solving the introduced uncertainty-guided/driven per-view depth optimization. We observed a convincing improvement in the reconstruction accuracy using Eq.(3) (refer main paper) optimization that combine surface normal and depth map (Fig.~\ref{fig:refined_depth_results_bear} \textbf{Right}).

\section{Discussion}
\formattedparagraph{\textit{(i)} Additional benefits of the proposed mobile robotic MVPS.} 
In addition to the benefits mentioned in the paper, robotic MVPS brings the advantage of applying the MVPS idea to online, incremental, and active 3D acquisition settings that can also help us avoid occlusion by moving the robot to an apt location for acquisition. Moreover, robotic MVPS allows precise control over an object's image acquisition and modalities. Our approach allows for more surface geometry capture if required, capturing fine details, which may not be possible with a static setup; hence, it is a scalable and reusable MVPS solution.

\smallskip
\noindent
\formattedparagraph{\textit{(ii)} Key fusion idea is more of a classical one than a neural network model.} 
In this work, our focus is to propose a ``portable mobile robot vision-based automation system for MVPS and make it work on a limited hardware budget via an online incremental strategy''. So, we were bound not to get too sophisticated with modeling \& formulating our fusion idea. Approaches such as \citep{menini2021real, sandstrom2022learning} can be used for a sophisticated fusion pipeline. Yet, \citep{menini2021real, sandstrom2022learning} has a high memory requirement, hence not apt for our MVPS system.

{\small
\bibliographystyle{elsarticle-num}
\bibliography{main.bib}

\begin{thebibliography}{10}
\expandafter\ifx\csname url\endcsname\relax
  \def\url#1{\texttt{#1}}\fi
\expandafter\ifx\csname urlprefix\endcsname\relax\def\urlprefix{URL }\fi
\expandafter\ifx\csname href\endcsname\relax
  \def\href#1#2{#2} \def\path#1{#1}\fi

\bibitem{kaya2023multi}
B.~Kaya, S.~Kumar, C.~Oliveira, V.~Ferrari, L.~Van~Gool, Multi-view photometric stereo revisited, in: Proceedings of the IEEE/CVF Winter Conference on Applications of Computer Vision, 2023, pp. 3126--3135.

\bibitem{kaya2022uncertainty}
B.~Kaya, S.~Kumar, C.~Oliveira, V.~Ferrari, L.~Van~Gool, Uncertainty-aware deep multi-view photometric stereo, in: Proceedings of the IEEE/CVF Conference on Computer Vision and Pattern Recognition, 2022, pp. 12601--12611.

\bibitem{li2020multi}
M.~Li, Z.~Zhou, Z.~Wu, B.~Shi, C.~Diao, P.~Tan, Multi-view photometric stereo: A robust solution and benchmark dataset for spatially varying isotropic materials, IEEE Transactions on Image Processing 29 (2020) 4159--4173.

\bibitem{schoenberger2016sfm}
J.~L. Schonberger, J.-M. Frahm, Structure-from-motion revisited, in: Proceedings of the IEEE/CVF Conference on Computer Vision and Pattern Recognition, IEEE, 2016, pp. 4104--4113.

\bibitem{furukawa2015multi}
Y.~Furukawa, C.~Hern{\'a}ndez, Multi-view stereo: A tutorial, Foundations and Trends{\textregistered} in Computer Graphics and Vision 9~(1-2) (2015) 1--148.

\bibitem{woodham1980photometric}
R.~J. Woodham, Photometric method for determining surface orientation from multiple images, Optical engineering 19~(1) (1980) 139--144.

\bibitem{hernandez2008multiview}
C.~Hernandez, G.~Vogiatzis, R.~Cipolla, Multiview photometric stereo, IEEE Transactions on Pattern Analysis and Machine Intelligence 30~(3) (2008) 548--554.

\bibitem{mildenhall2020nerf}
B.~Mildenhall, P.~P. Srinivasan, M.~Tancik, J.~T. Barron, R.~Ramamoorthi, R.~Ng, Nerf: Representing scenes as neural radiance fields for view synthesis, in: European conference on computer vision, Springer, 2020, pp. 405--421.

\bibitem{kerbl3Dgaussians}
B.~Kerbl, G.~Kopanas, T.~Leimk{\"u}hler, G.~Drettakis, 3d gaussian splatting for real-time radiance field rendering, ACM Transactions on Graphics 42~(4) (July 2023).

\bibitem{mueller2022instant}
T.~M\"uller, A.~Evans, C.~Schied, A.~Keller, Instant neural graphics primitives with a multiresolution hash encoding, ACM Trans. Graph. 41~(4) (2022) 102:1--102:15.

\bibitem{jain2024learning}
N.~Jain, S.~Kumar, L.~Van~Gool, Learning robust multi-scale representation for neural radiance fields from unposed images, International Journal of Computer Vision 132~(4) (2024) 1310--1335.

\bibitem{nehab2005efficiently}
D.~Nehab, S.~Rusinkiewicz, J.~Davis, R.~Ramamoorthi, Efficiently combining positions and normals for precise 3d geometry, ACM Transactions on Graphics (Proc. of ACM SIGGRAPH 2005) 24~(3) (2005) 536--543.

\bibitem{moons20093d}
T.~Moons, L.~Van~Gool, M.~Vergauwen, 3D reconstruction from multiple images: principles, Now Publishers Inc, 2009.

\bibitem{chatterjee2015photometric}
A.~Chatterjee, V.~Madhav~Govindu, Photometric refinement of depth maps for multi-albedo objects, in: Proceedings of the IEEE Conference on Computer Vision and Pattern Recognition, 2015, pp. 933--941.

\bibitem{bronstein2008numerical}
A.~M. Bronstein, M.~M. Bronstein, R.~Kimmel, Numerical geometry of non-rigid shapes, Springer Science \& Business Media, 2008.

\bibitem{kaya2021neural}
B.~Kaya, S.~Kumar, F.~Sarno, V.~Ferrari, L.~Van~Gool, Neural radiance fields approach to deep multi-view photometric stereo, in: Proceedings of the IEEE/CVF Winter Conference on Applications of Computer Vision, 2022, pp. 1965--1977.

\bibitem{park2016robust}
J.~Park, S.~N. Sinha, Y.~Matsushita, Y.-W. Tai, I.~S. Kweon, Robust multiview photometric stereo using planar mesh parameterization, IEEE transactions on pattern analysis and machine intelligence 39~(8) (2016) 1591--1604.

\bibitem{ren2011pocket}
P.~Ren, J.~Wang, J.~Snyder, X.~Tong, B.~Guo, Pocket reflectometry, ACM Transactions on Graphics (TOG) 30~(4) (2011) 1--10.

\bibitem{dong2010manifold}
Y.~Dong, J.~Wang, X.~Tong, J.~Snyder, Y.~Lan, M.~Ben-Ezra, B.~Guo, Manifold bootstrapping for svbrdf capture, ACM Transactions on Graphics (TOG) 29~(4) (2010) 1--10.

\bibitem{zhang2022iron}
K.~Zhang, F.~Luan, Z.~Li, N.~Snavely, Iron: Inverse rendering by optimizing neural sdfs and materials from photometric images, in: Proceedings of the IEEE/CVF Conference on Computer Vision and Pattern Recognition, 2022, pp. 5565--5574.

\bibitem{zhao2023mvpsnet}
D.~Zhao, D.~Lichy, P.-N. Perrin, J.-M. Frahm, S.~Sengupta, Mvpsnet: Fast generalizable multi-view photometric stereo, in: Proceedings of the IEEE/CVF International Conference on Computer Vision, 2023, pp. 12525--12536.

\bibitem{lichy2021shape}
D.~Lichy, J.~Wu, S.~Sengupta, D.~W. Jacobs, Shape and material capture at home, in: Proceedings of the IEEE/CVF Conference on Computer Vision and Pattern Recognition, 2021, pp. 6123--6133.

\bibitem{Cheng_2023_CVPR}
Z.~Cheng, J.~Li, H.~Li, Wildlight: In-the-wild inverse rendering with a flashlight, in: Proceedings of the IEEE/CVF Conference on Computer Vision and Pattern Recognition (CVPR), 2023, pp. 4305--4314.

\bibitem{wang2021neus}
P.~Wang, L.~Liu, Y.~Liu, C.~Theobalt, T.~Komura, W.~Wang, Neus: Learning neural implicit surfaces by volume rendering for multi-view reconstruction, Advances in Neural Information Processing Systems 34 (2021) 27171--27183.

\bibitem{shi2016benchmark}
B.~Shi, Z.~Wu, Z.~Mo, D.~Duan, S.-K. Yeung, P.~Tan, A benchmark dataset and evaluation for non-lambertian and uncalibrated photometric stereo, in: Proceedings of the IEEE Conference on Computer Vision and Pattern Recognition, 2016, pp. 3707--3716.

\bibitem{chen2019self}
G.~Chen, K.~Han, B.~Shi, Y.~Matsushita, K.-Y.~K. Wong, Self-calibrating deep photometric stereo networks, in: Proceedings of the IEEE Conference on Computer Vision and Pattern Recognition, 2019, pp. 8739--8747.

\bibitem{kaya2021uncalibrated}
B.~Kaya, S.~Kumar, C.~Oliveira, V.~Ferrari, L.~Van~Gool, Uncalibrated neural inverse rendering for photometric stereo of general surfaces, in: Proceedings of the IEEE/CVF Conference on Computer Vision and Pattern Recognition, 2021, pp. 3804--3814.

\bibitem{ikehata2018cnn}
S.~Ikehata, Cnn-ps: Cnn-based photometric stereo for general non-convex surfaces, in: Proceedings of the European Conference on Computer Vision (ECCV), 2018, pp. 3--18.

\bibitem{gal2015bayesian}
Y.~Gal, Z.~Ghahramani, Bayesian convolutional neural networks with {B}ernoulli approximate variational inference, in: 4th International Conference on Learning Representations (ICLR) workshop track, 2016.

\bibitem{ranftl2020towards}
R.~Ranftl, K.~Lasinger, D.~Hafner, K.~Schindler, V.~Koltun, Towards robust monocular depth estimation: Mixing datasets for zero-shot cross-dataset transfer, IEEE transactions on pattern analysis and machine intelligence 44~(3) (2020) 1623--1637.

\bibitem{ranftl2021vision}
R.~Ranftl, A.~Bochkovskiy, V.~Koltun, Vision transformers for dense prediction, in: Proceedings of the IEEE/CVF international conference on computer vision, 2021, pp. 12179--12188.

\bibitem{birkl2023midas}
R.~Birkl, D.~Wofk, M.~M{\"u}ller, Midas v3.1 -- a model zoo for robust monocular relative depth estimation, arXiv preprint arXiv:2307.14460 (2023).

\bibitem{liu2023single}
C.~Liu, S.~Kumar, S.~Gu, R.~Timofte, L.~Van~Gool, Single image depth prediction made better: A multivariate gaussian take, in: Proceedings of the IEEE/CVF Conference on Computer Vision and Pattern Recognition, 2023, pp. 17346--17356.

\bibitem{liu2023va}
C.~Liu, S.~Kumar, S.~Gu, R.~Timofte, L.~Van~Gool, Va-depthnet: A variational approach to single image depth prediction, in: The Eleventh International Conference on Learning Representations (ICLR), 2023.

\bibitem{shi2014photometric}
B.~Shi, K.~Inose, Y.~Matsushita, P.~Tan, S.-K. Yeung, K.~Ikeuchi, Photometric stereo using internet images, in: 2014 2nd International Conference on 3D Vision, Vol.~1, IEEE, 2014, pp. 361--368.

\bibitem{antensteiner2018variational}
D.~Antensteiner, S.~Stolc, T.~Pock, Variational depth and normal fusion algorithms for 3d reconstruction, in: Woman in Computer Vision Workshop (WiCV)@ Computer Vision and Pattern Recognition (CVPR) 2018, 2018.

\bibitem{hager2013limited}
W.~W. Hager, H.~Zhang, The limited memory conjugate gradient method, SIAM Journal on Optimization 23~(4) (2013) 2150--2168.

\bibitem{liu1989limited}
D.~C. Liu, J.~Nocedal, On the limited memory bfgs method for large scale optimization, Mathematical programming 45~(1-3) (1989) 503--528.

\bibitem{berahas2016multi}
A.~S. Berahas, J.~Nocedal, M.~Tak{\'a}c, A multi-batch l-bfgs method for machine learning, Advances in Neural Information Processing Systems 29 (2016).

\bibitem{curless1996volumetric}
B.~Curless, M.~Levoy, A volumetric method for building complex models from range images, in: Proceedings of the 23rd annual conference on Computer graphics and interactive techniques, 1996, pp. 303--312.

\bibitem{yang2020teaser}
H.~Yang, J.~Shi, L.~Carlone, Teaser: Fast and certifiable point cloud registration, IEEE Transactions on Robotics 37~(2) (2020) 314--333.

\bibitem{chen2018ps}
G.~Chen, K.~Han, K.-Y.~K. Wong, Ps-fcn: A flexible learning framework for photometric stereo, in: Proceedings of the European Conference on Computer Vision (ECCV), 2018, pp. 3--18.

\bibitem{kingma2014adam}
D.~P. Kingma, J.~L. Ba, Adam: A method for stochastic gradient descent, in: ICLR: international conference on learning representations, ICLR US., 2015, pp. 1--15.

\bibitem{lorensen1987marching}
W.~E. Lorensen, H.~E. Cline, Marching cubes: A high resolution 3d surface construction algorithm, ACM siggraph computer graphics 21~(4) (1987) 163--169.

\bibitem{yao2018mvsnet}
Y.~Yao, Z.~Luo, S.~Li, T.~Fang, L.~Quan, Mvsnet: Depth inference for unstructured multi-view stereo, in: Proceedings of the European Conference on Computer Vision (ECCV), 2018, pp. 767--783.

\bibitem{wang2021patchmatchnet}
F.~Wang, S.~Galliani, C.~Vogel, P.~Speciale, M.~Pollefeys, Patchmatchnet: Learned multi-view patchmatch stereo, in: Proceedings of the IEEE/CVF Conference on Computer Vision and Pattern Recognition, 2021, pp. 14194--14203.

\bibitem{indelman2015incremental}
V.~Indelman, R.~Roberts, F.~Dellaert, Incremental light bundle adjustment for structure from motion and robotics, Robotics and Autonomous Systems 70 (2015) 63--82.

\bibitem{knapitsch2017tanks}
A.~Knapitsch, J.~Park, Q.-Y. Zhou, V.~Koltun, Tanks and temples: Benchmarking large-scale scene reconstruction, ACM Transactions on Graphics (ToG) 36~(4) (2017) 1--13.

\bibitem{lee2019monocular}
J.-H. Lee, C.-S. Kim, Monocular depth estimation using relative depth maps, in: Proceedings of the IEEE/CVF Conference on Computer Vision and Pattern Recognition, 2019, pp. 9729--9738.

\bibitem{kumar2019superpixel}
S.~Kumar, Y.~Dai, H.~Li, Superpixel soup: Monocular dense 3d reconstruction of a complex dynamic scene, IEEE transactions on pattern analysis and machine intelligence 43~(5) (2019) 1705--1717.

\bibitem{pulli1999multiview}
K.~Pulli, Multiview registration for large data sets, in: Second international conference on 3-d digital imaging and modeling (cat. no. pr00062), IEEE, 1999, pp. 160--168.

\bibitem{kirillov2023segany}
A.~Kirillov, E.~Mintun, N.~Ravi, H.~Mao, C.~Rolland, L.~Gustafson, T.~Xiao, S.~Whitehead, A.~C. Berg, W.-Y. Lo, et~al., Segment anything, in: Proceedings of the IEEE/CVF International Conference on Computer Vision, 2023, pp. 4015--4026.

\bibitem{noah22bringing}
N.~Z. Rothenberger, Bringing multi-view photometric stereo and mobile robotics together, MS Thesis, Computer Vision Lab, ETH Zurich (2022).

\bibitem{menini2021real}
D.~Menini, S.~Kumar, M.~R. Oswald, E.~Sandstr{\"o}m, C.~Sminchisescu, L.~Van~Gool, A real-time online learning framework for joint 3d reconstruction and semantic segmentation of indoor scenes, IEEE Robotics and Automation Letters 7~(2) (2021) 1332--1339.

\bibitem{sandstrom2022learning}
E.~Sandstr{\"o}m, M.~R. Oswald, S.~Kumar, S.~Weder, F.~Yu, C.~Sminchisescu, L.~Van~Gool, Learning online multi-sensor depth fusion, in: European Conference on Computer Vision, Springer, 2022, pp. 87--105.

\end{thebibliography}
}
\end{document}